\documentclass[journal]{IEEEtran}

\usepackage{times}
\usepackage{epsfig}
\usepackage{graphicx}
\usepackage{amsmath}
\usepackage{amssymb}

\usepackage{caption}
\usepackage{subcaption}
\usepackage{epsfig}
\usepackage{amsmath}
\usepackage{amssymb}
\usepackage{booktabs}
\usepackage{rotating}
\usepackage{algorithm} 
\usepackage{algpseudocode}
\usepackage{multirow}
\usepackage{color}

\usepackage{enumitem}
\usepackage{enumerate}

\newcommand{\etal}{\textit{et al}.}
\newcommand{\ie}{\textit{i}.\textit{e}.}
\newcommand{\eg}{\textit{e}.\textit{g}.}

\newif\ifshowcomments
\showcommentstrue

\ifshowcomments
\newcommand{\TODO}[1]{{\color{red}{[TODO: #1]}}}
\newcommand{\revised}[1]{{\color[rgb]{0.0,0.0,0.0}{#1}}}
\newcommand{\tywang}[1]{{\color[rgb]{0.7,0.7,0}{#1}}}
\newcommand{\phil}[1]{{\color[rgb]{0.9,0.1,0.1}{#1}}}

\else
\newcommand{\TODO}[1]{}
\newcommand{\revised}[1]{}
\newcommand{\tywang}[1]{}
\newcommand{\phil}[1]{}
\fi

\ifCLASSINFOpdf
\else
\fi
\usepackage[pagebackref=false,breaklinks=true,letterpaper=true,colorlinks,citecolor=black,linkcolor=black,bookmarks=false]{hyperref}

\begin{document}
%
%
\title{Revisiting Shadow Detection: \\ A New Benchmark Dataset for Complex World}
%
%

\author{Xiaowei~Hu, Tianyu~Wang, Chi-Wing~Fu,~\IEEEmembership{Member,~IEEE,} \\ Yitong Jiang,   Qiong Wang, 
	and~Pheng-Ann~Heng,~\IEEEmembership{Senior Member,~IEEE}
	\IEEEcompsocitemizethanks{
		\IEEEcompsocthanksitem X. Hu, T. Wang, C.-W. Fu, and Y. Jiang are with the Department of Computer Science and Engineering, The Chinese University of Hong Kong, Hong Kong SAR, China.
		\IEEEcompsocthanksitem Q. Wang is with Shenzhen Key Laboratory of Virtual Reality and Human Interaction Technology, Shenzhen Institutes of Advanced Technology, Chinese Academy of Sciences, Shenzhen 518055, China.
		\IEEEcompsocthanksitem P.-A. Heng is with the Department of Computer Science and Engineering, The Chinese University of Hong Kong, Hong Kong SAR, China and also with Guangdong-Hong Kong-Macao Joint Laboratory of Human-Machine Intelligence-Synergy Systems, Shenzhen Institutes of Advanced Technology, Chinese Academy of Sciences, Shenzhen, 518055, China.
		\IEEEcompsocthanksitem Corresponding author: Xiaowei Hu
	}
}


%
%

\markboth{IEEE Transactions on Image Processing}%
{Shell \MakeLowercase{\textit{et al.}}: Bare Demo of IEEEtran.cls for IEEE Journals}
%



\maketitle

\begin{abstract}
Shadow detection in general photos is a nontrivial problem, due to the complexity of the real world.
Though recent shadow detectors have already achieved remarkable performance on various benchmark data, their performance is still limited for general real-world situations.
In this work, we collected shadow images for multiple scenarios and compiled a new dataset of 10,500 shadow images, each with labeled ground-truth mask, for supporting shadow detection in the complex world.
Our dataset covers a rich variety of scene categories, with diverse shadow sizes, locations, contrasts, and types.
Further, we comprehensively analyze the complexity of the dataset, 
present a fast shadow detection network with a detail enhancement module to harvest shadow details, 
and demonstrate the effectiveness of our method to detect shadows in general situations.
\end{abstract}

\begin{IEEEkeywords}
Shadow detection, benchmark dataset, complex, deep neural network.
\end{IEEEkeywords}

%
\IEEEpeerreviewmaketitle



\section{Introduction}
\label{sec:introduction}

Shadows are formed in the 3D volume behind the objects that occlude the light.
The appearance of a shadow generally depends not only on the shape of the occluding object, but also on the light that shines on the object (its direction and strength) and the geometry of the background object on which the shadow is cast.
Yet, in general real photos, it is likely that we can observe multiple shadows cast by multiple objects and lights, while the shadows may lie on or go across multiple background objects in the scene.
Hence, shadow detection can be a very complicated problem in general situations.

From the research literature of computer vision and image processing~\cite{chondagar2015review,xu2006shadow}, 
the presence of shadows could degrade the performance of many object recognition tasks, e.g., object detection and tracking~\cite{cucchiara2003detecting,nadimi2004physical}, person re-identification~\cite{bekele2018implementing}, etc.
%
Also, the knowledge of shadows in a scene can help to estimate the light conditions~\cite{lalonde2009estimating,panagopoulos2011illumination} and scene geometry~\cite{karsch2011rendering,okabe2009attached}.
Thus, shadow detection has long been a fundamental problem.

\begin{figure*}[t!]
\centering
\includegraphics[width=0.97\linewidth]{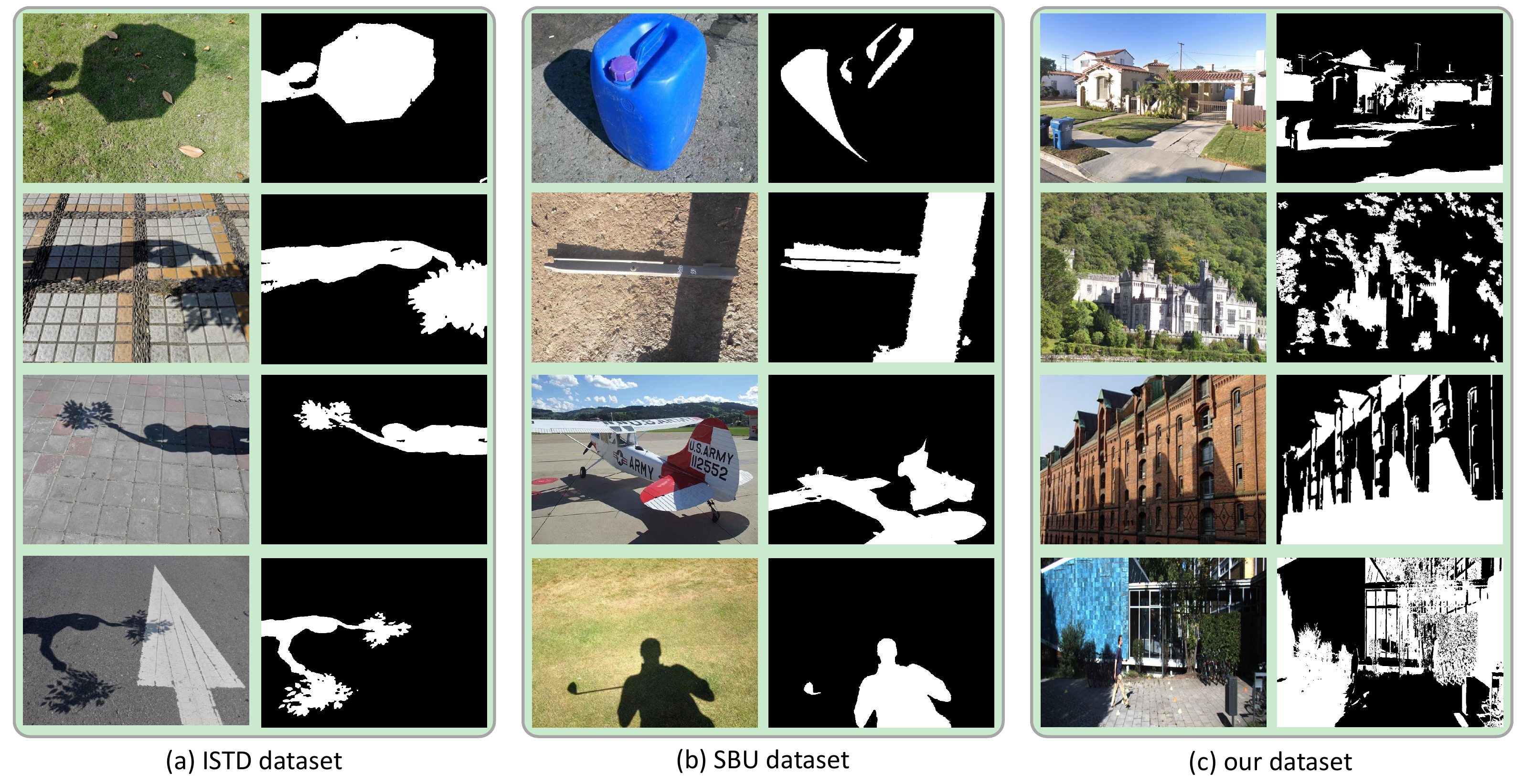}
\vspace*{-1.5mm}
\caption{Example shadow images and masks in ISTD~\cite{wang2018stacked}, SBU~\cite{hou2019large, vicente2016noisy, vicente2016large}, and our CUHK-Shadow dataset.}
\label{fig:dataset_comp}
\vspace*{-2mm}
\end{figure*}

At present, the de facto approach to detect shadows~\cite{vicente2016large,khan2016automatic,Hu_2018_CVPR,le2018a+d,wang2018stacked,ding2019argan,hu2019direction,zheng2019distraction, hou2019large} is based on deep neural networks, which have demonstrated notable performance on various benchmark data~\cite{guo2011single, hou2019large, vicente2016noisy, vicente2016large, wang2018stacked, zhu2010learning}.
However, existing datasets contain mainly shadows cast by {\em single or few separate objects\/}.
They do not adequately model the complexity of shadows in the real world; see Figures~\ref{fig:dataset_comp} (a) \& (b).
Though recent methods~\cite{zheng2019distraction,zhu2018bidirectional} already achieved nearly-saturated performance on the benchmarks with a balanced error rate (BER) less than 4\% on the SBU~\cite{vicente2016noisy, vicente2016large, hou2019large} and ISTD~\cite{wang2018stacked} datasets, if we use them to detect shadows in {\em various types of real-world situations\/}, their performance is rather limited; see Section~\ref{sec:experiments}.
Also, current datasets 
contain {\em mainly cast shadows with few self shadows\/}, thus limiting the shadow detection performance in general situations.
Note that when an object occludes the light and casts shadows, self shadows are regions on the object that do not receive direct light, while cast shadows are projections of the object on some other background objects.

%
In this work, we prepare a new dataset to support shadow detection in complex real-world situations.
Our dataset contains 10,500 shadow images, each with labeled ground-truth mask. 
Apart from the dataset size, it has three main advantages when comparing with the existing data.
First, the shadow images are collected from {\em diverse scenes\/},~\eg, cities, buildings, satellite maps, and roads, which are 
general and challenging situations that existing data do not exhibit.
%
Second, our dataset includes cast shadows on background objects and also {\em self shadows\/} on occluding objects.
%
%
Third, besides the training and testing sets, our dataset provides {\em a validation set\/} for tuning the training parameters and performing ablation study for deep models.
This helps to reduce the risk of overfitting.
%

Besides, we design a fast shadow detection network called FSDNet by adopting the direction-aware spatial context module~\cite{hu2019direction,Hu_2018_CVPR} to aggregate global information from high-level feature maps and formulating a detail enhancement module to harvest shadow details in low-level feature maps.
Also, we perform a comprehensive statistical analysis on our dataset to study its complexity, and evaluate the performance of various shadow detectors and FSDNet on the data.
Experimental results show that FSDNet performs favorably against the state-of-the-arts; particularly, it has only 4M model parameters, so it can achieve real-time performance, while detecting shadow with good quality.
\revised{\emph{The dataset, evaluation code, and source code are publicly available at \url{https://xw-hu.github.io/}.}}

\section{Related Work}
\label{sec:related}

Shadow detection on single image has been widely studied in computer vision research.
%
Early methods focus on illumination models or machine learning algorithms by exploring various hand-crafted shadow features,~\eg, geometrical properties~\cite{salvador2004cast,panagopoulos2011illumination}, spectrum ratios~\cite{tian2016new}, color~\cite{lalonde2010detecting,vicente2015leave,guo2011single,vicente2018leave}, texture~\cite{zhu2010learning,vicente2015leave,guo2011single,vicente2018leave}, edge~\cite{lalonde2010detecting,zhu2010learning,huang2011characterizes}, and T-junction~\cite{lalonde2010detecting}. 
These features, however, have limited capability to distinguish between the shadow and non-shadow regions, so approaches based on them often fail to detect shadows in general real-world environments.

Later, methods based on features learned from deep convolutional neural networks (CNNs) demonstrate remarkable improved performance on various benchmarks, especially when large training data is available.
Khan~\etal~\cite{khan2014automatic} adopt CNNs to learn features at the super-pixel level and object boundaries, and use a conditional random field to predict shadow contours.
Shen~\etal~\cite{shen2015shadow} predict the structures of shadow edges from a structured CNN and adopt a global shadow optimization for shadow recovery.
Vicente~\etal~\cite{vicente2016large} train a stacked-CNN to detect shadows by recovering the noisy shadow annotations.
%
Nguyen~\etal~\cite{nguyen2017shadow} introduce a sensitive parameter to 
the loss in a conditional generative adversarial network to solve the unbalanced labels of shadow and non-shadow regions.

Hu~\etal~\cite{Hu_2018_CVPR,hu2019direction} 
aggregate global context features via two rounds of data translations and formulate the direction-aware spatial context features to detect shadows.
Wang~\etal~\cite{wang2018stacked} jointly detect and remove shadows by stacking two conditional generative adversarial networks. 
Le~\etal~\cite{le2018a+d} adopt a shadow attenuation network to generate adversarial training samples, further for training a shadow detection network. 
Zhu~\etal~\cite{zhu2018bidirectional} formulate a recurrent attention residual module to selectively use the global and local features in a bidirectional feature pyramid network.
Zheng~\etal~\cite{zheng2019distraction} present a distraction-aware shadow detection network by explicitly revising the false negative and false positive regions found by other shadow detection methods.
Ding~\etal~\cite{ding2019argan} jointly detect and remove shadows in a recurrent manner.
While these methods have achieved high accuracy in detecting shadows in current benchmarks~\cite{hou2019large, vicente2016noisy, vicente2016large},
their performances are still limited for complex real environments; see 
experiments in Section~\ref{sec:experiments_SOTA}.
%
Apart from shadow detection, recent works explore deep learning methods to remove shadows~\cite{khan2016automatic,qu2017deshadownet,hu2019mask,ding2019argan,Le_2019_ICCV}, to generate shadows~\cite{liu2020arshadowgan}, and to detect the shadow-object associations~\cite{wang2020instance}, but the training data for these tasks also contains mainly shadows cast by a few objects.

\begin{figure*}[t!]
	\centering
	\includegraphics[width=0.99\linewidth]{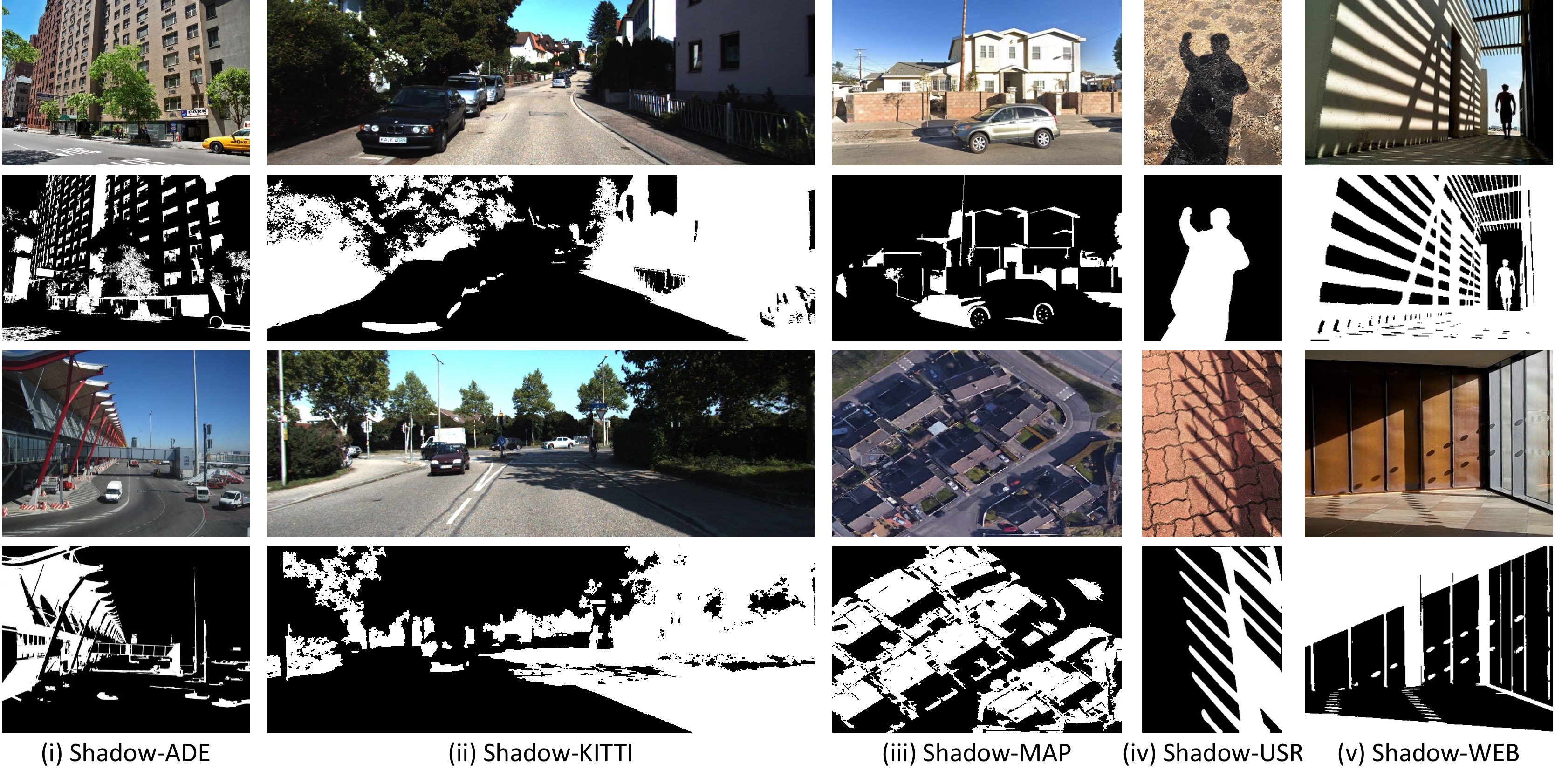}
	\vspace*{-1.5mm}
	\caption{Example shadow images and shadow masks for categories (i) to (v) in our dataset; see Section~\ref{sec:dataset_construction} for details.}
	\label{fig:our_data}
	\vspace*{-1mm}
\end{figure*}

%


\section{Our CUHK-Shadow Dataset}
\label{sec:dataset}

Existing datasets for shadow detection,~\ie, UCF~\cite{zhu2010learning}, UIUC~\cite{guo2011single}, SBU~\cite{hou2019large, vicente2016noisy, vicente2016large}, and ISTD~\cite{wang2018stacked} have been widely used in the past decade. 
Among them, pioneering ones,~\ie, UCF and UIUC, contain only 245 and 108 images, respectively, so deep models trained on them certainly have limited generalization capability, as shown in~\cite{vicente2016noisy, vicente2016large}.
For the more recent ones, SBU has 4,087 training images and 638 testing images, whereas ISTD has 1,330 training and 540 testing triples of shadow images, shadow-free images, and shadow masks.
Typically, ISTD has only 135 background scenes; while SBU features a wider variety of scenes, both datasets provide mainly shadows cast by single or a few objects.
In contrast, our dataset, named as CUHK-Shadow, contains 10,500 shadow images, each with mask, featuring shadows in diverse situations; see Figure~\ref{fig:dataset_comp} for example images randomly picked from ISTD, SBU, and our new dataset.
%


\subsection{Building the Dataset}
\label{sec:dataset_construction}
%

To start, we collected shadow images from five different sources by ourselves:
(i) {\em Shadow-ADE\/}: 1,132 images from the ADE20K dataset~\cite{zhou2017scene,zhou2019semantic} with shadows cast mainly by buildings;
(ii) {\em Shadow-KITTI\/}: 2,773 images from the KITTI dataset~\cite{geiger2012we}, featuring shadows of vehicles, trees, and objects along the roads;
(iii) {\em Shadow-MAP\/}: 1,595 remote-sensing and street-view photos from Google Map;
(iv) {\em Shadow-USR\/}: 2,445 images from the USR dataset~\cite{hu2019mask} with mainly people and object shadows; and
(v) {\em Shadow-WEB\/}: 2,555 Internet images found by using a web crawler on Flickr website with keywords ``shadow'' and ``shadow building.''
Hence, the dataset enables us to assess the shadow detection performance for various scenarios, corresponding to the need for different applications. Also, images in different categories have domain differences, so our dataset has the potential to be used to evaluate and guide domain adaptation methods for shadow detection.


Next, we hired a professional company to label the shadows in these images.
%
First, we randomly selected some sample images for the company to label, and discussed their labeled masks with them on the label quality.
\revised{Later, they (around 30 persons) labeled all the dataset images, then we (three of the paper authors) examined the results one by one and asked the company (around five persons) to re-label the low-quality ones.
Next, we checked and refined all the masks ourselves.
It took around five months for the data labeling, five days for examining the masks, and two weeks for refining the masks.}

Further, we followed COCO~\cite{lin2014microsoft} to analyze the mask quality by comparing the results with the masks labeled manually by ourselves.
In detail, we randomly selected 100 images from the dataset, adopted more expensive operations (iPad Pro and Apple Pencil) to label the masks by three of the paper authors, and used the majority vote to obtain more accurate ground truths.
We then compute how close the data labels are with respect to the manual annotations prepared by ourselves; the matching percentage is found to be $96.46\%$.
%
\revised{Note that, since data labeling involves subjective judgement, we argue that it is not always possible to obtain perfect masks through manual annotations. 
In the future, we shall explore more robust learning approaches to detect shadows with less dependence on the labels.}
Figure~\ref{fig:our_data} shows example shadow images and masks for the five categories of shadow images in our CUHK-Shadow dataset.
We randomly split the images in each category into a training set, validation set, and testing set with a ratio of 7:1:2.
So, we have 7,350 training images, 1,050 validation images, and 2,100 testing images in total.
To the best of our knowledge, CUHK-Shadow is currently the largest shadow detection dataset with labeled shadow masks.
Also, it is the first shadow detection dataset with a validation set, and features a wide variety of real-world situations.


\begin{figure*}[tp]
\centering
\includegraphics[width=0.99\linewidth]{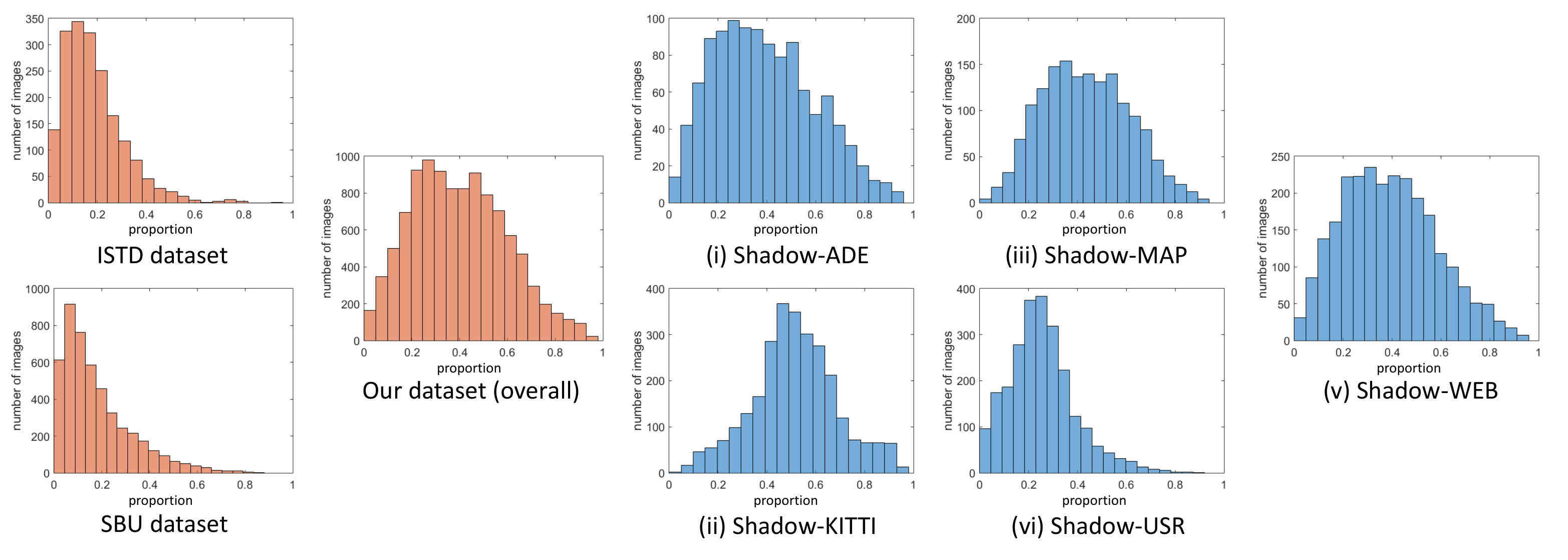}
\vspace{-2mm}
\caption{Analysis on the shadow area proportion for different datasets.
	Shadows in the ISTD and SBU datasets have mainly small shadows, while our CUHK-Shadow has more diverse types of shadows with wider ranges of sizes in the shadow images.}
\label{fig:shadow_area}
\vspace*{-1mm}
\end{figure*}


\subsection{Dataset Complexity}

To provide a comprehensive understanding of the dataset, we performed a series of statistical analysis on the shadow images and compared the statistical results with the ISTD and SBU datasets in the following aspects.
%


\vspace{1.5mm}
\noindent
\textbf{Shadow area proportion.} \
First, we find the proportion of pixels (range: [0,1]) occupied by shadows in each image.
Figure~\ref{fig:shadow_area} (left) shows the histogram plots of shadow area proportion for ISTD, SBU, and our CUHK-Shadow dataset.
%
From the histograms, we can see that most images in ISTD and SBU have relatively small shadow regions, while our CUHK-Shadow has more diverse shadow areas compared with them.
%
%
%
%
Figures~\ref{fig:shadow_area} (right) further reports histogram plots for the five scene categories in CUHK-Shadow.
Interestingly, the distribution in {\em Shadow-WEB\/} can serve as a good reference for general real-world shadows, since the images were obtained from the Internet, while {\em Shadow-KITTI\/} features mainly shadows for road images and {\em Shadow-USR\/} features mainly shadows for people and objects, so the shadow areas in these images have less variation than other categories.
%
%


\begin{table}  [tp]
	\begin{center}
		\caption{Number of separated shadow regions per image
		in ISTD~\cite{wang2018stacked}, SBU~\cite{hou2019large, vicente2016noisy, vicente2016large}, and our CUHK-Shadow.}
		\vspace*{-2mm}
		\label{table:individual_shadow_num}
		\resizebox{0.75\linewidth}{!}{%
			\begin{tabular}{cc|c|l}
				\multicolumn{2}{c}{\multirow{2}{*}{}} & \multicolumn{2}{|c}{Count} \\ \cline{3-4} 
				\multicolumn{2}{c}{} & \multicolumn{1}{|l|}{mean} & std \\ \hline
				\multicolumn{2}{c|}{ISTD} & 1.51 & 1.14 \\ \hline
				\multicolumn{2}{c|}{SBU} & 3.44 & 3.02 \\ \hline
				\multicolumn{1}{c|}{\multirow{6}{*}{\begin{tabular}[c]{@{}c@{}}Our\\ CUHK-Shadow\\ Dataset\end{tabular}}} & Overall & 7.52 & 7.04 \\ \cline{2-4} 
				\multicolumn{1}{c|}{} & Shadow-ADE & 10.11 & 7.98 \\ \cline{2-4} 
				\multicolumn{1}{c|}{} & Shadow-KITTI & 9.63 & 5.92 \\ \cline{2-4} 
				\multicolumn{1}{c|}{} & Shadow-MAP & 8.94 & 6.91 \\ \cline{2-4} 
				\multicolumn{1}{c|}{} & Shadow-USR & 4.31 & 6.36\\ \cline{2-4} 
				\multicolumn{1}{c|}{} & Shadow-WEB & 6.26 & 6.96\\
			\end{tabular}%
		}
	\end{center}
	\vspace{-3mm}
\end{table}

\vspace{1.5mm}
\noindent
\textbf{Number of shadows per image.} \
Next, we group connected shadow pixels and count the number of separated shadows per image in ISTD, SBU, and CUHK-Shadow.
%
%
To avoid influence of noisy labels, we ignore shadow regions whose area is less than 0.05\% of the whole image.
Table~\ref{table:individual_shadow_num} reports the resulting statistics, showing that ISTD and SBU only have around 1.51 and 3.44 shadow regions per image, while CUHK-Shadow has far more shadow regions per image on average.
This certainly reveals the complexity of CUHK-Shadow.
Note also that there are more than ten separate shadow regions per image in {\em Shadow-ADE\/}, showing the challenge of detecting shadows in this data category in CUHK-Shadow.


\begin{figure}[t!]
\centering
\includegraphics[width=0.95\linewidth]{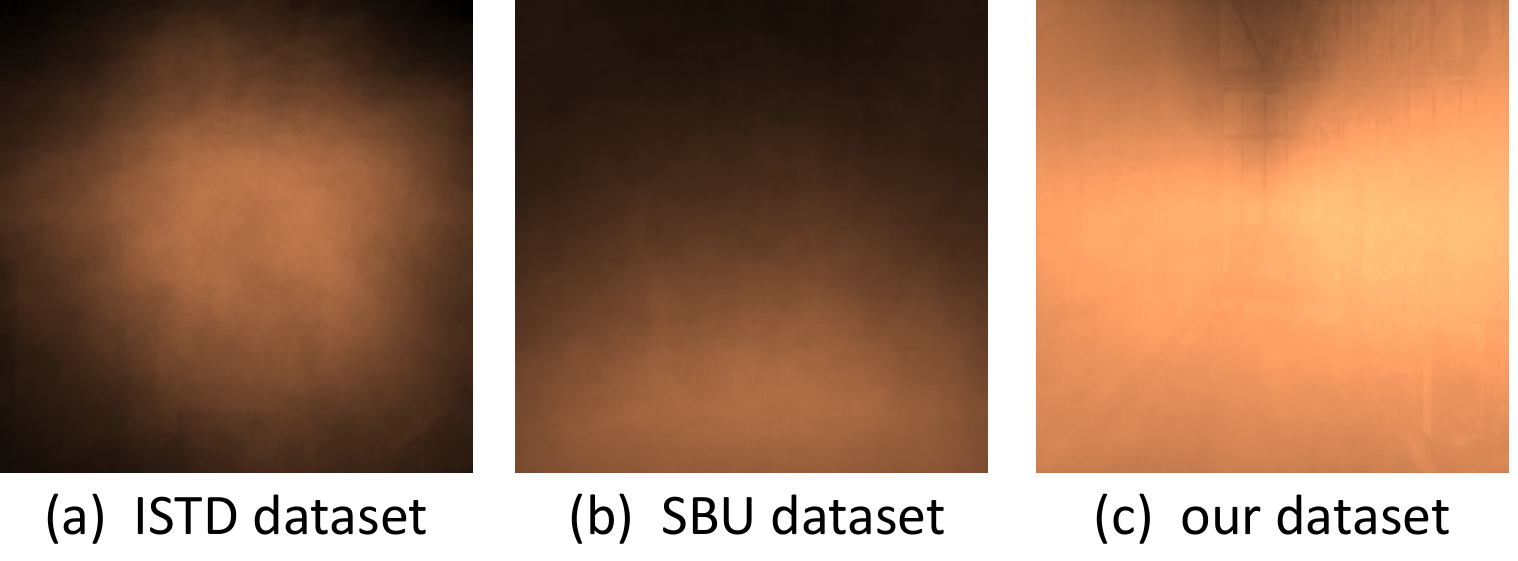}
\vspace*{-1mm}
\caption{Shadow location distributions.
	Lighter (darker) colors indicate larger (smaller) chances of having shadows.}
\label{fig:location}
\end{figure}

\vspace{1.5mm}
\noindent
\textbf{Shadow location distribution.} \
Further, we study shadow locations in image space by resizing all shadow masks to 512$\times$512 and summing them up per dataset.
So, we can obtain a per-pixel probability value about shadow occurrence.
Figure~\ref{fig:location} shows the results for the three datasets, revealing that shadows in CUHK-Shadow cover a wider spatial range, except for the top regions, which are often related to the sky.
In contrast, shadows in ISTD locate mainly in the middle, while shadows in SBU locate mainly on the bottom.

\begin{figure}[t!]
	\centering
	\includegraphics[width=0.6\linewidth]{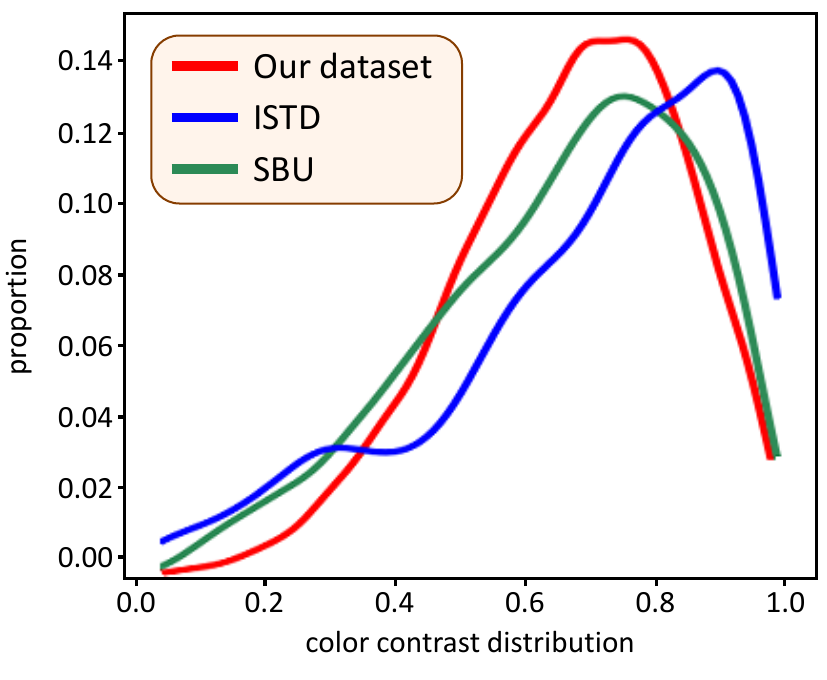}
	\vspace{-2mm}
	\caption{Color contrast distributions of different datasets.}
	\label{fig:color_intensity}
\end{figure}

\vspace{1.5mm}
\noindent
\textbf{Color contrast distribution.} \
Real-world shadows are often more soft instead of being entirely dark.
This means that the color contrast in shadow and non-shadow regions may not be high.
Here, we follow~\cite{li2014secrets,Yang_2019_ICCV} to measure the ${\chi}^2$ distances between the color histograms of the shadow and non-shadow regions in each image.
Figure~\ref{fig:color_intensity} plots the color contrast distribution for images in the three datasets, where a contrast value (horizontal axis in the plot) of one means high color contrast, and vice versa.
From the results, the color contrast in CUHK-Shadow is lower than both ISTD and SBU (except a few images with extremely low contrast), so it is more challenging to detect shadows in CUHK-Shadow.


\begin{figure*}[!tp]
	\centering
	\includegraphics[width=0.98\linewidth]{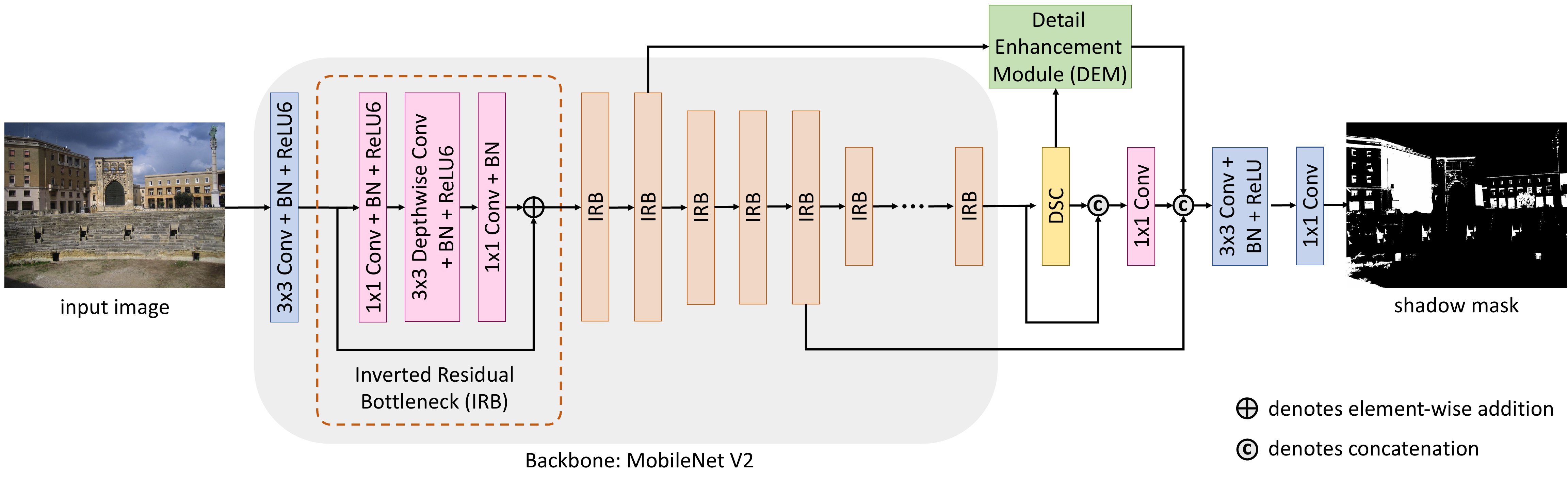}
	\caption{Illustration of our fast shadow detection network (FSDNet).
		Note that the height of the boxes indicates the size of the associated feature maps.
		BN and IRB denote batch normalization and inverted residual bottleneck, respectively.}
	%
	%
	\label{fig:architecture}
\end{figure*}

\vspace{1.5mm}
\noindent
\textbf{Shadow detection performance.} \
Last, to reveal the complexity of CUHK-Shadow with respect to ISTD and SBU, we compare the performance of two recent shadow detectors, DSDNet~\cite{zheng2019distraction} and BDRAR~\cite{zhu2018bidirectional}, on the three datasets.
DSDNet and BDRAR achieve 2.17 and 2.69 balanced error rate (BER) on ISTD, 3.45 and 3.64 BER on SBU, but only 8.09 and 9.13 BER on CUHK-Shadow, respectively, showing that it is more challenging to detect shadows in CUHK-Shadow.


\vspace{1.5mm}
\noindent
\textbf{Summary.} \
Overall, CUHK-Shadow not only contains far more shadow images and covers a rich variety of scene categories, but also features more diverse proportion of shadow areas, separated shadows, and shadow locations, as well as lower color contrast between shadows and non-shadows.
Having said that, it also means that CUHK-Shadow is more challenging and complex for shadow detection.
This is also evidenced by the experimental results to be shown in Table~\ref{table:compare_with_SOTA}.




\subsection{Evaluation Metrics}
\label{sec:eval}

Balanced error rate (BER)~\cite{vicente2016large} is a common metric to evaluate shadow detection performance, where shadow and non-shadow regions contribute equally to the overall performance without considering their relative areas:
\begin{equation}
BER \ = \ (1-\frac{1}{2}(\frac{TP}{TP+FN}+\frac{TN}{TN+FP}))\times 100 \ ,
\end{equation}
where $TP$, $TN$, $FP$ and $FN$ are true positives, true negatives, false positives, and false negatives, respectively.
To compute these values, we have to first quantize the predicted shadow mask into a binary mask, then
compare this binary mask with the ground truth mask.
A lower BER value indicates a better detection result.


The recent deep neural networks~\cite{Hu_2018_CVPR,zhu2018bidirectional,wang2018stacked,le2018a+d,zheng2019distraction,ding2019argan} predict shadow masks in continuous values, which indicate the probability of a pixel of being inside a shadow. 
However, BER is designed for evaluating binary predictions, in which predicted values less than $0.5$ are set as zeros, or ones, otherwise.
Therefore, some false predictions with gray values (false negatives with predicted values greater than $0.5$ or false positives with predicted values less than $0.5$) give no effects to the qualitative results, but they do affect the visual results; see Section~\ref{sec:experiments_SOTA} for the comparisons.
%
\revised{To evaluate continuous predictions, we introduce the $F_\beta^\omega$-measure~\cite{margolin2014evaluate}, which extends $TP$, $TN$, $FP$, and $FN$ to
%
%
%
\begin{eqnarray}
\label{eq:Fb1}
TP^\omega
&=&
(1-E^\omega) \odot \ G \ ,
\nonumber
\\
TN^\omega
&=&
(1-E^\omega) \odot \ (1-G) \ , 
\nonumber
\\
FP^\omega
&=&
E^\omega \odot \ (1-G) \ , \ 
\text{and} \ \
\nonumber
\\
FN^\omega
&=&
E^\omega \odot \ G \ ,
\end{eqnarray}
where
$G$ is the ground truth image,
$\odot$ is the element-wise multiplication,
%
and $E^{\omega}$ denotes the weighted error map:
%
\begin{equation}
E^{\omega} = min(E, \ E\mathbb{A}) \odot \mathbb{B} \ ,
\end{equation}
where $E = |G-M|$, $M$ is the predicted shadow mask, $min$ computes the minimum value for each element, and $\mathbb{A}$ captures the dependency between pairs of foreground pixels:
\begin{equation}
\mathbb{A}(i,j) = 
\left\{
\begin{array}{ll}
   \frac{1}{\sqrt{2\pi \sigma^2}} e^{-\frac{d(i,j)^2}{2\sigma^2}}  &\forall i,j, \  G(i)=1, G(j)=1 \ ;  \\
   1 \  &\forall i,j, \  G(i)=0, i=j \ ; \\
   0 \  &otherwise \ ,
\end{array}
\right.
\end{equation}
where $d(i,j)$ is the Euclidean distance between pixels $i$ and $j$, and we followed~\cite{margolin2014evaluate} to set $\sigma^2 = 5$. 
Also, $\mathbb{B}$ represents the varying importance to the false detections based on their distance from the foreground:
\begin{equation}
\mathbb{B}(i) = 
\left\{
\begin{array}{ll}
1  &\forall i, \  G(i)=1 \ ;  \\
2-e^{\alpha \Delta(i)} \  &otherwise \ ,
\end{array}
\right.
\end{equation}
where $\Delta(i) = \min\limits_{G(j)=1}d(i,j)$ and $\alpha = \frac{1}{5}\ln(0.5)$. 
Finally, the weighted precision, weighted recall, and $F_\beta^\omega$ are defined as:
\begin{eqnarray}
\text{precision}^\omega = \frac{TP^\omega}{TP^\omega+FP^\omega} \ ,  \ \text{recall}^\omega = \frac{TP^\omega}{TP^\omega+FN^\omega} \ ,
\end{eqnarray}
\begin{equation}
F_\beta^\omega \ = \ (1+\beta^2) \frac{\text{precision}^\omega \cdot \text{recall}^\omega}{\beta^2 \cdot \text{precision}^\omega + \text{recall}^\omega} \ ,
\end{equation}
and we followed~\cite{margolin2014evaluate} to set $\beta^2$ as one.
Overall, a larger $F_\beta^\omega$ indicates a better result.
\emph{The dataset and evaluation code are publicly available at \url{https://xw-hu.github.io/}.} 
}



\section{Methodology}
\label{sec:method}

\begin{table*}  [tp]
	\begin{center}
		\caption{Ablation study on the validation set of our dataset.}
		\label{table:ablation}
  	\resizebox{0.98\linewidth}{!}{%
			\begin{tabular}{c|c|c|c|c|c|c|c|c|c|c|c|c}
				\hline
			    \multirow{2}{*}{Method} & \multicolumn{2}{c|}{Overall} & \multicolumn{2}{c|}{shadow-ADE}& \multicolumn{2}{c|}{shadow-KITTI} & \multicolumn{2}{c|}{shadow-MAP} & \multicolumn{2}{c|}{shadow-USR} & \multicolumn{2}{c}{shadow-WEB}  \\
			    
			    \cline{2-13}
			    & $F_\beta^\omega$ & BER & $F_\beta^\omega$ & BER & $F_\beta^\omega$ & BER & $F_\beta^\omega$ & BER & $F_\beta^\omega$ & BER & $F_\beta^\omega$ & BER \\
	
				\hline
				\hline
			
				basic & 84.32 & 10.94 & 74.81 & 13.92 & 87.05 & 9.35 & 82.51 & 11.15 & 88.78 & 5.17 & 82.43 & 10.72  \\
				
				
				basic+DSC & 84.84 & 10.74 & 75.52 & 13.77 & 87.30 & 9.10 &82.81 & 10.90 & 89.90 & 4.90 & 82.71 & 10.28 \\


				
				\hline
				
				FSDNet w/o DEM  & 85.88 & 9.78 & 76.58 &  12.74 & 88.59 & 7.89 & 84.61 & 9.54 & 89.70 &4.95 & 84.17 & 9.42 \\
				
				
				\hline
				
				FSDNet-high-only & 84.70  & 10.54 & 75.52 & 13.34 & 87.15 & 9.15 & 82.75 & 10.92  & 89.44 & 4.90 & 82.79 & 10.15 \\ 
				
				
				\hline

				FSDNet (our full pipeline) & \textbf{{86.12}} & \textbf{{9.58}} & \textbf{{76.85}} & \textbf{{12.49}} & \textbf{{88.70}} & \textbf{{7.82}} & \textbf{{84.86}} & \textbf{{9.29}} & \textbf{{90.00}} & \textbf{{4.62}} & \textbf{{84.50}} & \textbf{{8.98}} \\
				
				\hline
				
			\end{tabular} }
	\end{center}
	\vspace{-2mm}
\end{table*}



\noindent
\textbf{Network architecture.} \ 
Figure~\ref{fig:architecture} shows the overall architecture of our fast shadow detection network (FSDNet).
It takes a shadow image as input and outputs a shadow mask in an end-to-end manner.
First, we use MobileNet V2~\cite{sandler2018mobilenetv2} as the backbone with a series of inverted residual bottlenecks (IRBs) to extract feature maps in multiple scales.
%
Each IRB contains a 1$\times$1 convolution, a 3$\times$3 depthwise convolution~\cite{chollet2017xception}, and another 1$\times$1 convolution, with a skip connection to add the input and output feature maps.
Also, it adopts batch normalization~\cite{ioffe2015batch} after each convolution and ReLU6~\cite{howard2017mobilenets} after the first two convolutions.
%
Second, we employ the direction-aware spatial context (DSC) module~\cite{hu2019direction,Hu_2018_CVPR} after the last convolutional layer of the backbone to harvest the DSC features, which contain global context information for recognizing shadows.
%

Third, low-level feature maps of the backbone contain rich fine details that can help discover shadow boundaries and tiny shadows.
%
So, we further formulate the detail enhancement module (DEM) by harvesting shadow details in low-level feature maps when the distance between the DSC feature and low-level feature is large.
%
%
Last, we concatenate the DEM-refined low-level feature, mid-level feature, and high-level feature, then use a series of convolution layers to predict the output shadow mask; see Figure~\ref{fig:architecture} for details.



\begin{figure}[!tp]
	\centering
	\includegraphics[width=0.98\linewidth]{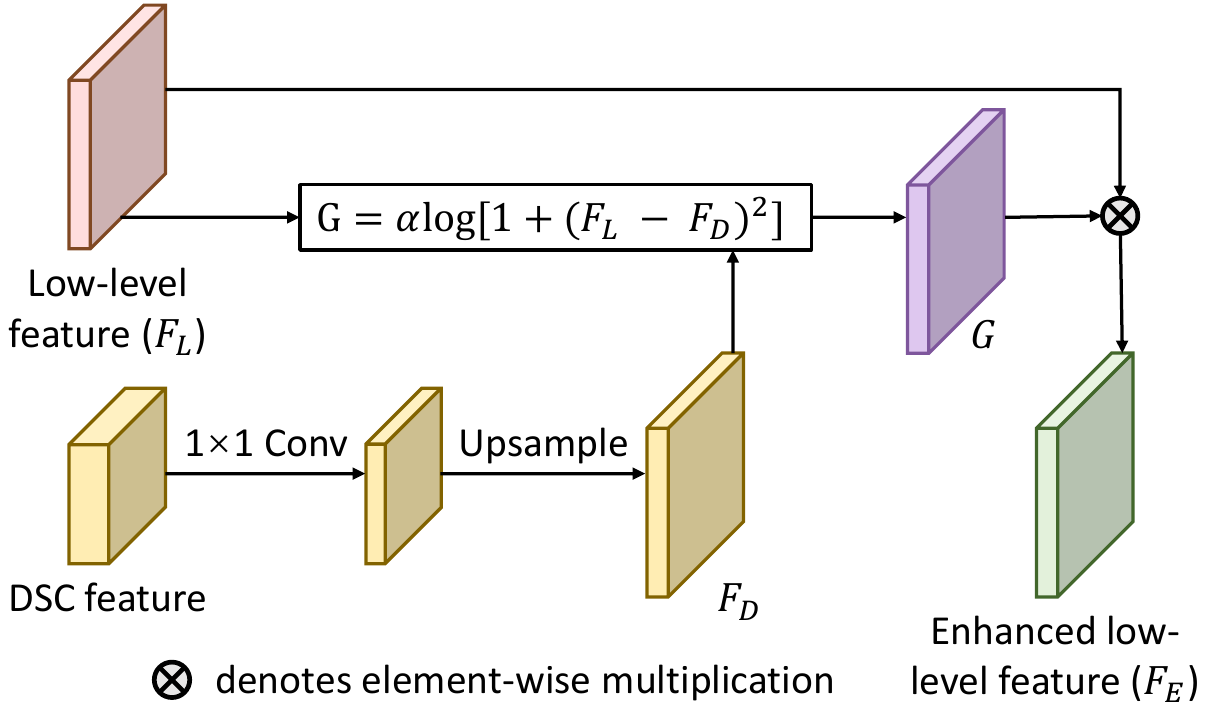}
	\caption{The detail enhancement module (DEM).}
	\label{fig:DEM}
\end{figure}


%
%
%

%
\begin{figure*}[t]
	\centering

         \ \\
  \vspace*{0.3mm}
  \begin{subfigure}{0.136\linewidth}
  	\includegraphics[width=\textwidth]{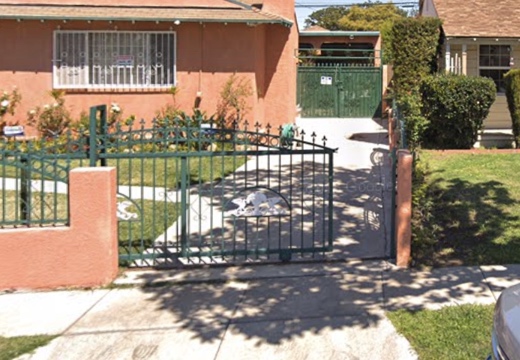}
  \end{subfigure}
  \begin{subfigure}{0.136\linewidth}
  	\includegraphics[width=\textwidth]{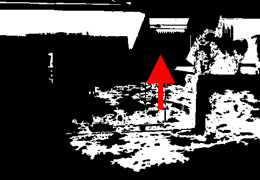}
  \end{subfigure}
  \begin{subfigure}{0.136\linewidth}
  	\includegraphics[width=\textwidth]{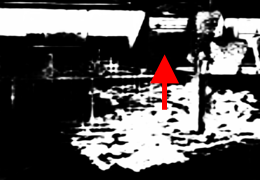}
  \end{subfigure}
  \begin{subfigure}{0.136\linewidth}
  	\includegraphics[width=\textwidth]{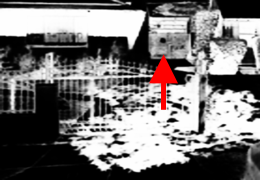}
  \end{subfigure}
  \begin{subfigure}{0.136\linewidth}
  	\includegraphics[width=\textwidth]{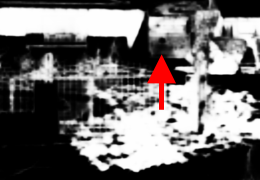}
  \end{subfigure}
  \begin{subfigure}{0.136\linewidth}
  	\includegraphics[width=\textwidth]{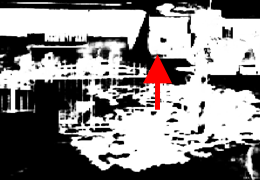}
   \end{subfigure}
   \begin{subfigure}{0.136\linewidth}
   	\includegraphics[width=\textwidth]{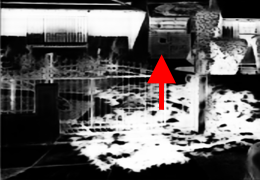}
   \end{subfigure}
	
	  \ \\

	\vspace*{0.3mm}
	\begin{subfigure}{0.136\linewidth}
		\includegraphics[width=\textwidth]{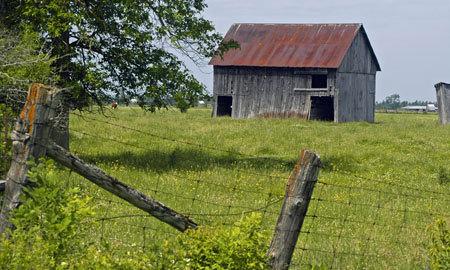}
	\end{subfigure}
	\begin{subfigure}{0.136\linewidth}
		\includegraphics[width=\textwidth]{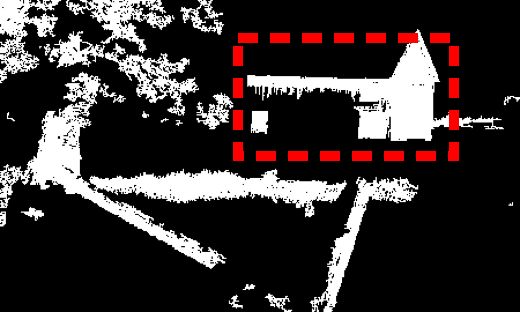}
	\end{subfigure}
	\begin{subfigure}{0.136\linewidth}
		\includegraphics[width=\textwidth]{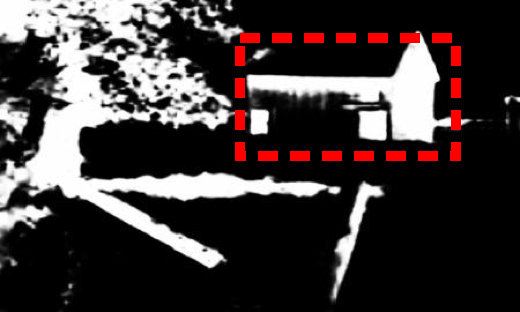}
	\end{subfigure}
	\begin{subfigure}{0.136\linewidth}
		\includegraphics[width=\textwidth]{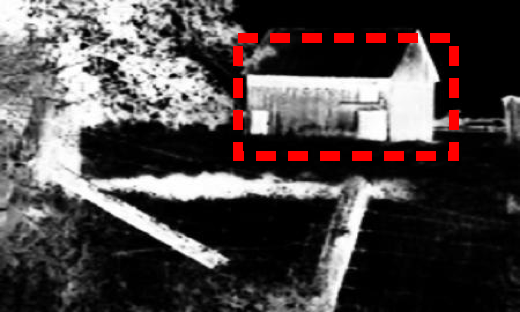}
	\end{subfigure}
	\begin{subfigure}{0.136\linewidth}
		\includegraphics[width=\textwidth]{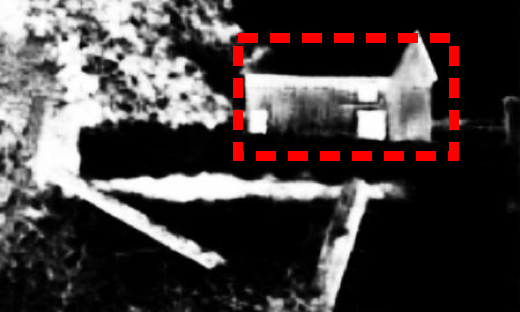}
	\end{subfigure}
	\begin{subfigure}{0.136\linewidth}
		\includegraphics[width=\textwidth]{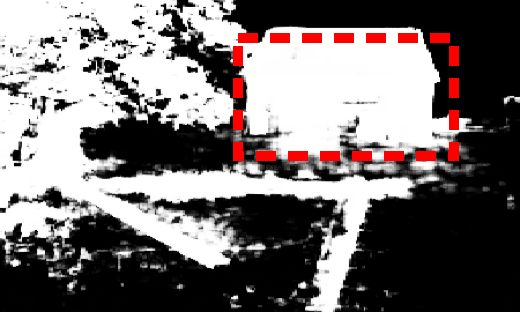}
	\end{subfigure}
	\begin{subfigure}{0.136\linewidth}
		\includegraphics[width=\textwidth]{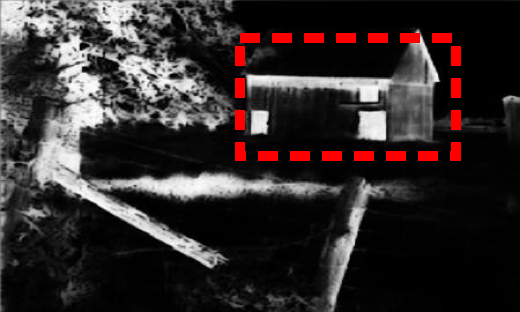}
	\end{subfigure}
	\ \\
	
	\vspace*{0.3mm}
	\begin{subfigure}{0.136\linewidth}
		\includegraphics[width=\textwidth]{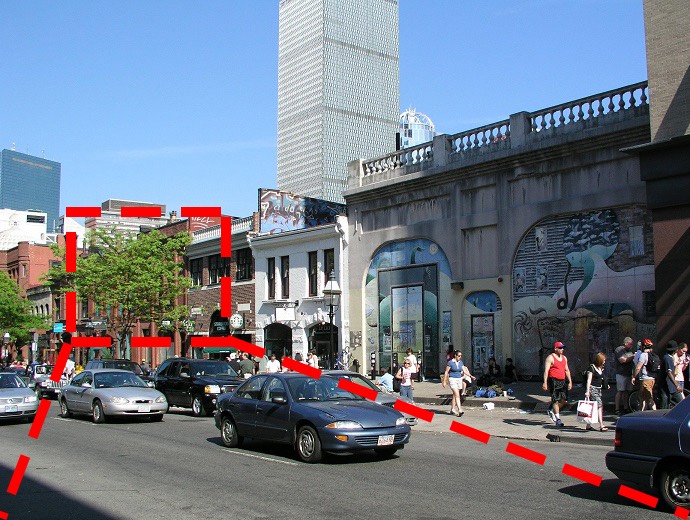}
		\vspace{-5.5mm} 
	\end{subfigure}
	\begin{subfigure}{0.136\linewidth}
		\includegraphics[width=\textwidth]{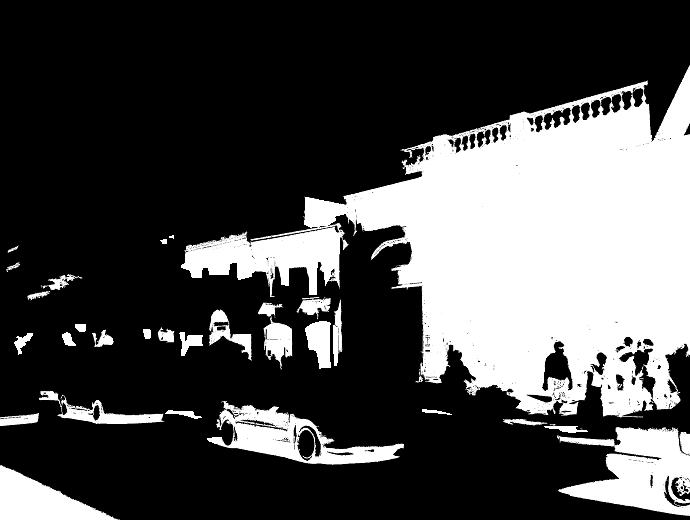}
		\vspace{-5.5mm} 
	\end{subfigure}
	\begin{subfigure}{0.136\linewidth}
		\includegraphics[width=\textwidth]{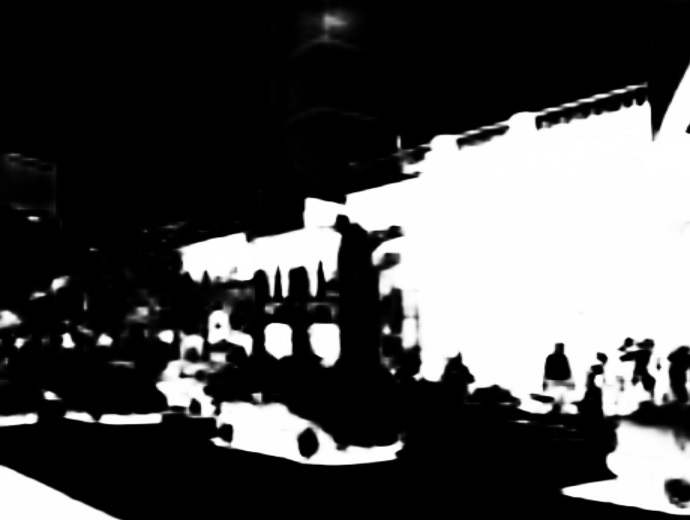}
		\vspace{-5.5mm} 
	\end{subfigure}
	\begin{subfigure}{0.136\linewidth}
		\includegraphics[width=\textwidth]{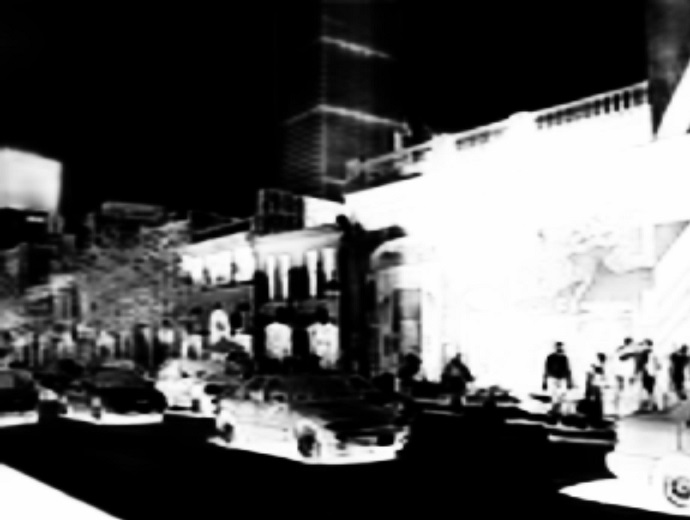}
		\vspace{-5.5mm} 
	\end{subfigure}
	\begin{subfigure}{0.136\linewidth}
		\includegraphics[width=\textwidth]{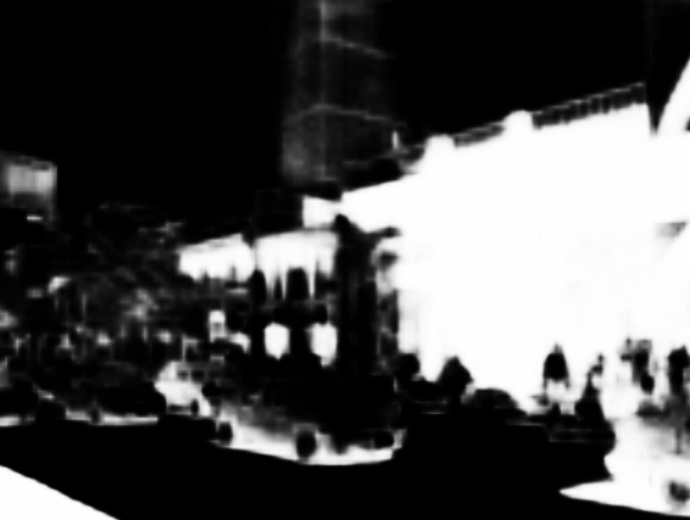}
		\vspace{-5.5mm} 
	\end{subfigure}
	\begin{subfigure}{0.136\linewidth}
		\includegraphics[width=\textwidth]{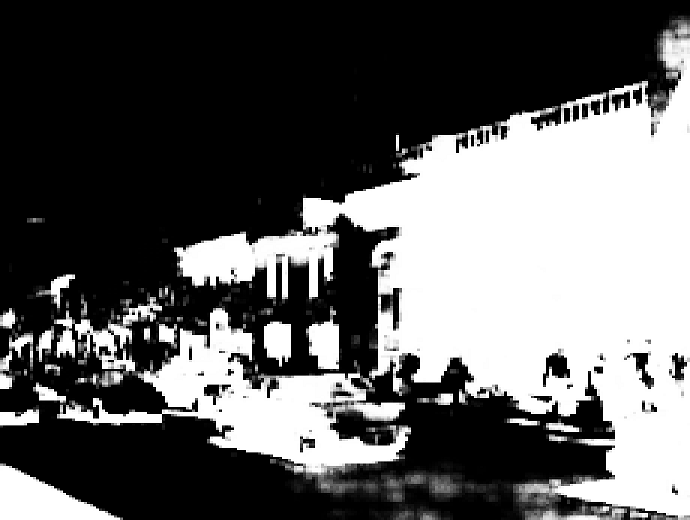}
		\vspace{-5.5mm} 
	\end{subfigure}
	\begin{subfigure}{0.136\linewidth}
		\includegraphics[width=\textwidth]{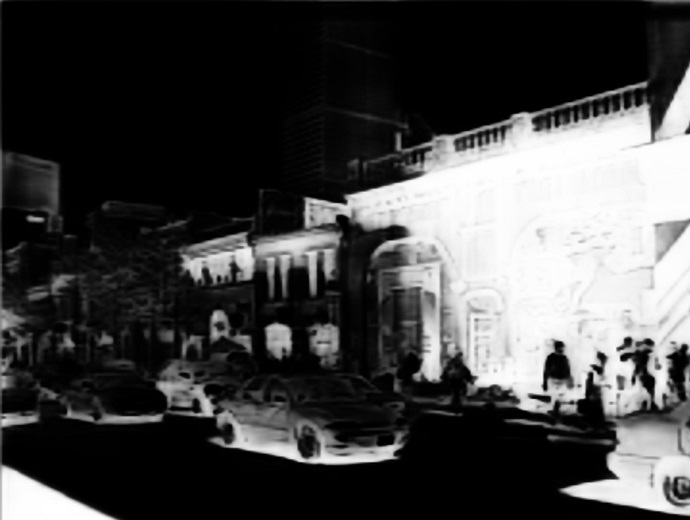}
		\vspace{-5.5mm} 
	\end{subfigure}
	
	\ \\

	\vspace*{1.5mm}
	\begin{subfigure}{0.136\linewidth}
		\includegraphics[width=\textwidth]{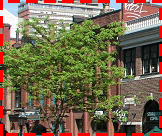}
		\vspace{-5.5mm} \caption{Input image}
	\end{subfigure}
	\begin{subfigure}{0.136\linewidth}
		\includegraphics[width=\textwidth]{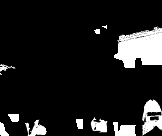}
		\vspace{-5.5mm} \caption{Ground truth}
	\end{subfigure}
	\begin{subfigure}{0.136\linewidth}
		\includegraphics[width=\textwidth]{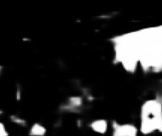}
		\vspace{-5.5mm} \caption{FSDNet (ours)}
	\end{subfigure}
	\begin{subfigure}{0.136\linewidth}
		\includegraphics[width=\textwidth]{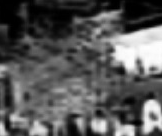}
		\vspace{-5.5mm} \caption{DSDNet~\cite{zheng2019distraction}}
	\end{subfigure}
	\begin{subfigure}{0.136\linewidth}
		\includegraphics[width=\textwidth]{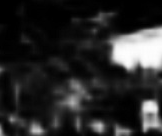}
		\vspace{-5.5mm} \caption{BDRAR~\cite{zhu2018bidirectional}}
	\end{subfigure}
	\begin{subfigure}{0.136\linewidth}
		\includegraphics[width=\textwidth]{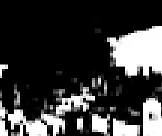}
		\vspace{-5.5mm} \caption{A+D Net~\cite{le2018a+d}}
	\end{subfigure}
	\begin{subfigure}{0.136\linewidth}
		\includegraphics[width=\textwidth]{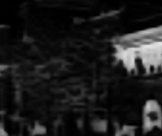}
		\vspace{-5.5mm} \caption{DSC~\cite{hu2019direction,Hu_2018_CVPR}}
	\end{subfigure}
	
	\caption{Visual comparison of the shadow masks produced by our method and by other shadow detection methods.}
	\label{fig:comparisons_visual}
    
\end{figure*}

\vspace{1.5mm}
\noindent
\textbf{Detail enhancement module.} \
Figure~\ref{fig:DEM} shows the structure of the detail enhancement module (DEM).
With low-level feature $F_{L}$ and DSC feature $F_D$ as inputs, it first reduces the number of feature channels of $F_D$ by a 1$\times$1 convolution and upsample it to the size of $F_{L}$.
Then, we compute gate map $G$ to measure the importance of the detail structures based on the distance between the DSC feature and low-level feature:
\begin{equation}
G \ = \ \alpha \ log( \ 1 \ + \ (F_L - F_D)^2 \ ) \ ,
\end{equation}
where $(F_L - F_D)^2$ reports the distance between the two features, which is rescaled by a logarithm function.
Then, we follow~\cite{fu2019adaptive} to introduce a learnable parameter $\alpha$ to adjust the scale of the gate map.
%
%
In the end, we multiply gate map $G$ with input low-level feature $F_L$ to enhance the spatial details and produce the refined low-level feature $F_E$.
Note that this module only introduces a few parameters (a 1$\times$1 convolution and parameter $\alpha$), so the computing time is negligible.



\vspace{1.5mm}
\noindent
\textbf{Training strategies.} \
We use the weights of MobileNet V2 trained on ImageNet~\cite{deng2009imagenet} for classification to initialize the backbone network parameters, and initialize the parameters in other layers by random noise.
We use stochastic gradient descent with momentum 0.9 and weight decay 0.0005 to optimize the network by minimizing the $L_1$ loss between the ground-truth and predicted shadow masks.
We set the initial learning rate as 0.005, reduce it by the poly strategy~\cite{liu2015parsenet} with a power of 0.9, and stop the learning after 50k iterations.
Last, we implement the network on PyTorch, train it on a GeForce GTX 1080 Ti GPU with a mini-batch size of six, and horizontally flip the images as data augmentation.


\section{Experimental Results}
\label{sec:experiments}


\begin{table*}  [t!]
	\begin{center}
		\caption{Comparing with state-of-the-art methods in terms of $F_\beta^\omega$ and BER.
		We trained all methods on our training set and tested them on our testing set without using any post-processing method such as CRF.
		Note that ``FPS'' stands for ``frames per second,'' which is evaluated on a single GeForce GTX 1080 Ti GPU with a batch size of one and image size of 512$\times$512.}
		\label{table:compare_with_SOTA}
		\resizebox{1.0\linewidth}{!}{%
			\begin{tabular}{c|c|c|c|c|c|c|c|c|c|c|c|c|c|c}
				\hline
			    \multirow{2}{*}{Method} & \multirow{2}{*}{FPS} & \multirow{2}{*}{Params (M)} & \multicolumn{2}{c|}{Overall} & \multicolumn{2}{c|}{shadow-ADE}& \multicolumn{2}{c|}{shadow-KITTI} & \multicolumn{2}{c|}{shadow-MAP} & \multicolumn{2}{c|}{shadow-USR} & \multicolumn{2}{c}{shadow-WEB}  \\
			    
			    \cline{4-15}
			    & & & $F_\beta^\omega$ & BER & $F_\beta^\omega$ & BER & $F_\beta^\omega$ & BER & $F_\beta^\omega$ & BER & $F_\beta^\omega$ & BER & $F_\beta^\omega$ & BER \\
			    
			    \hline
			    \hline
			    
			    FSDNet (ours)& \textbf{77.52} & \textbf{4.40} & \textbf{86.27} & 8.65 &
			    \textbf{79.04} & 10.14 &
			    \textbf{88.78} & 8.10 & 
			    \textbf{83.22} & 9.95 &
			    \textbf{90.86} & 4.40 & 
			    \textbf{84.27} & 9.75  \\

				DSDNet~\cite{zheng2019distraction}& 21.96 & 58.16 & 82.59 & \textbf{8.27} & 
				74.75 & \textbf{9.80} &
				86.52 &  \textbf{7.92} &
				78.56 & \textbf{9.59} &
				86.41 & \textbf{4.03} &
				80.64 & \textbf{9.20}  \\

				BDRAR~\cite{zhu2018bidirectional}& 18.20  & 42.46 & 83.51 & 9.18 &
				76.53 & 10.47 &
				85.35 & 9.73 &
				79.77 & 10.56 &
				88.56 & 4.88 &
				82.09 & 10.09  \\

				A+D Net~\cite{le2018a+d}& 68.02 & 54.41 & 83.04 & 12.43 & 
				73.64 & 15.89 &
				88.83 & 9.06 &
				78.46 & 13.72 & 
				86.63 & 6.78 &
				80.32 & 14.34  \\

				DSC~\cite{hu2019direction,Hu_2018_CVPR}& 4.95 & 79.03  & 82.76 & 8.65 &  
				75.26 & 10.49 &  
				87.03 & 7.58 & 
				79.26 & 9.56 & 
				85.34 & 4.53 & 
				81.16 & 9.92  \\
				
				\hline

				$R^3$Net~\cite{deng18r} &26.43 & 56.16&
				81.36 & 8.86 & 
				73.32 & 10.18 &
				84.95 & 8.20 &
				76.46 & 10.80 &
				86.03 & 4.97 &
				79.61 & 10.21 \\
				
				\hline
				
			    MirrorNet~\cite{Yang_2019_ICCV} & 16.01 & 127.77 &
				78.29 & 13.39 &
				69.83 & 15.20 & 
				79.92 & 12.77 &
				74.22 & 14.03 &
				85.12 & 7.08 &
				76.26 & 15.30\\

				\hline
				PSPNet~\cite{zhao2017pyramid} & 12.21& 65.47 &
				84.93 & 10.65 &
				76.76 & 12.38 &
				88.12 & 9.48 &
				81.14 & 12.65 &
				90.42 & 5.68 &
				82.20 & 12.75  \\
				
				
				\hline

			\end{tabular} }
	\end{center}
	\vspace{-4mm}
\end{table*}

 \begin{figure*}[h]
	\centering

    \vspace*{0.3mm}
    \begin{subfigure}{0.136\linewidth}
    	\includegraphics[width=\textwidth]{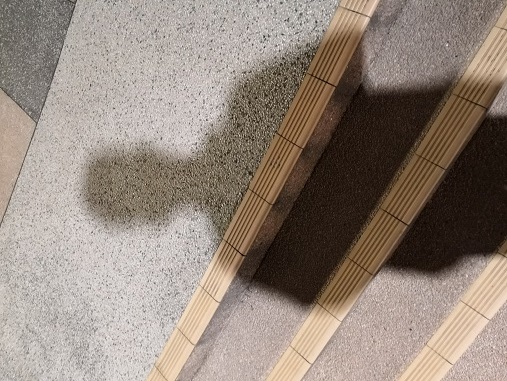}
    \end{subfigure}
    \begin{subfigure}{0.136\linewidth}
    	\includegraphics[width=\textwidth]{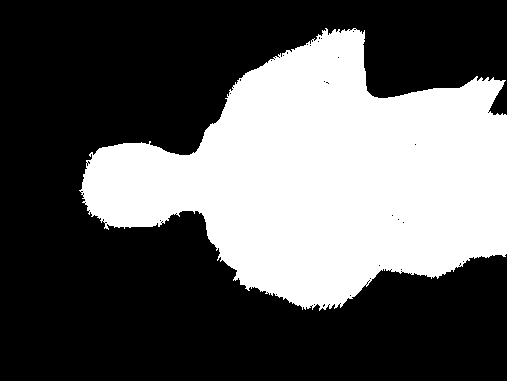}
    \end{subfigure}
    \begin{subfigure}{0.136\linewidth}
    	\includegraphics[width=\textwidth]{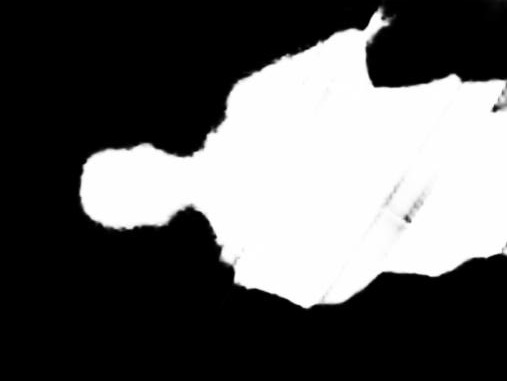}
    \end{subfigure}
    \begin{subfigure}{0.136\linewidth}
    	\includegraphics[width=\textwidth]{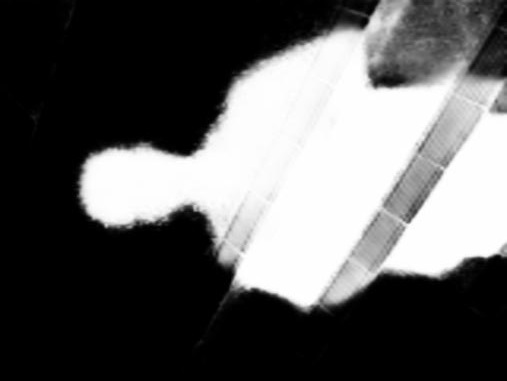}
    \end{subfigure}
    \begin{subfigure}{0.136\linewidth}
    	\includegraphics[width=\textwidth]{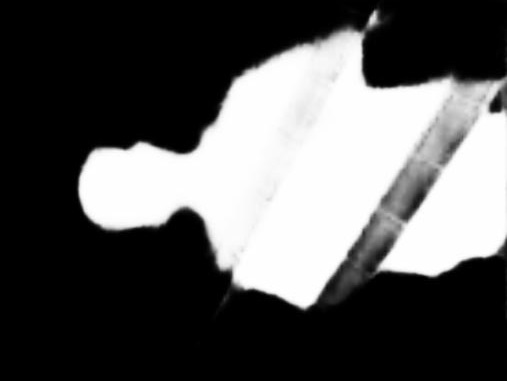}
    \end{subfigure}
    \begin{subfigure}{0.136\linewidth}
    	\includegraphics[width=\textwidth]{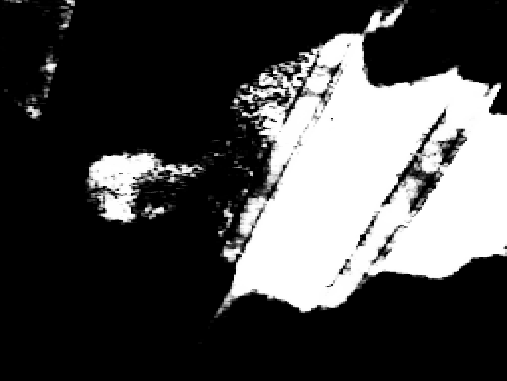}
    \end{subfigure}
    \begin{subfigure}{0.136\linewidth}
    	\includegraphics[width=\textwidth]{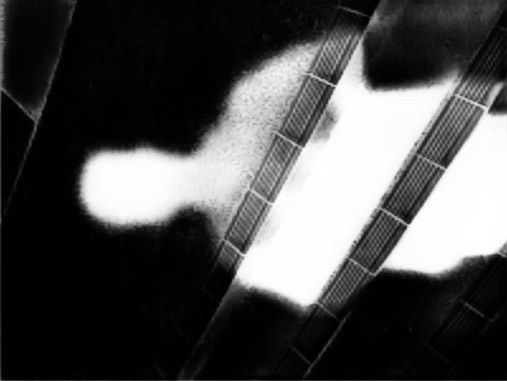}
    \end{subfigure}
    
       \ \\

       \vspace*{0.3mm}
  \begin{subfigure}{0.136\linewidth}
  	\includegraphics[width=\textwidth]{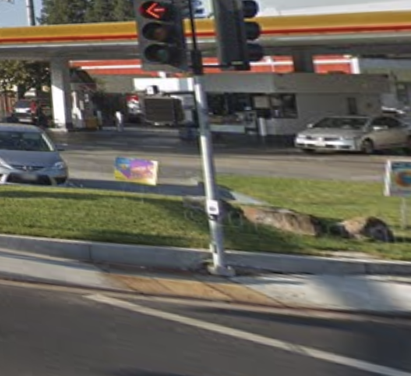}
  \end{subfigure}
  \begin{subfigure}{0.136\linewidth}
  	\includegraphics[width=\textwidth]{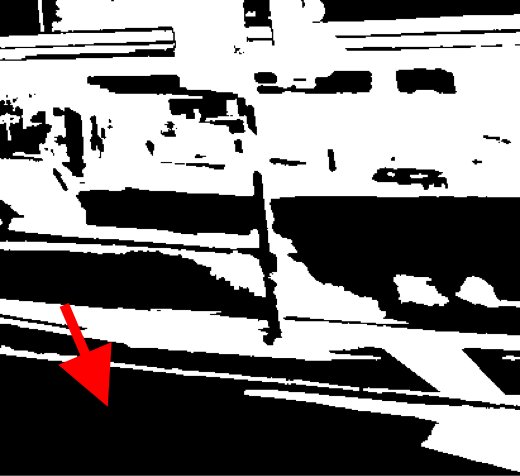}
  \end{subfigure}
  \begin{subfigure}{0.136\linewidth}
  	\includegraphics[width=\textwidth]{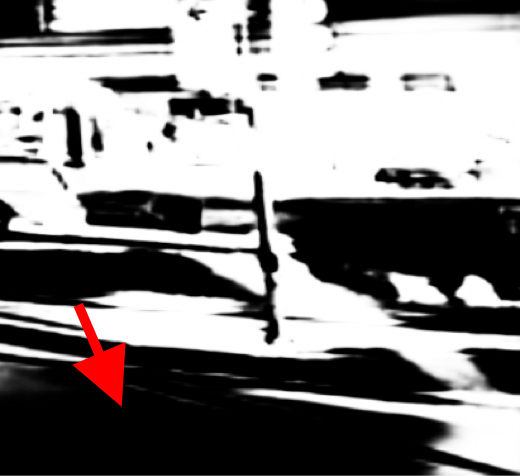}
  \end{subfigure}
  \begin{subfigure}{0.136\linewidth}
  	\includegraphics[width=\textwidth]{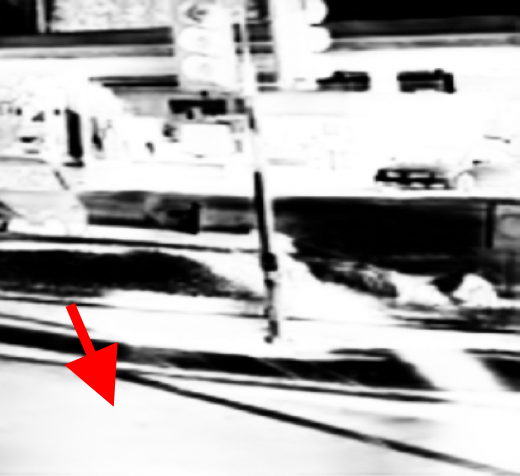}
  \end{subfigure}
  \begin{subfigure}{0.136\linewidth}
  	\includegraphics[width=\textwidth]{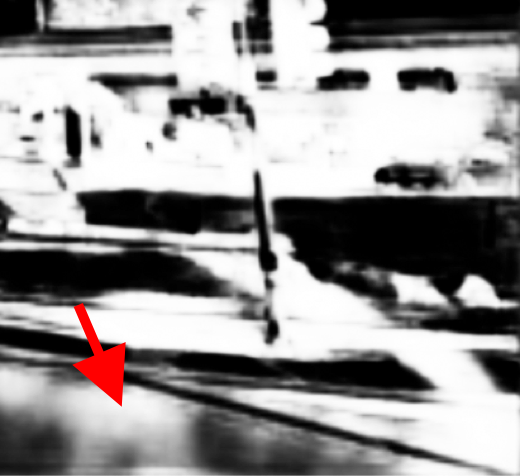}
  \end{subfigure}
  \begin{subfigure}{0.136\linewidth}
  	\includegraphics[width=\textwidth]{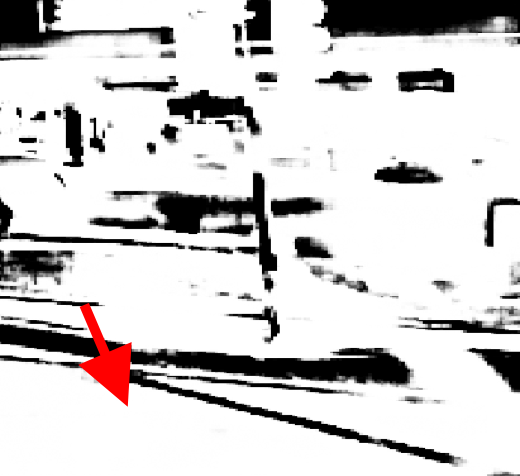}
  \end{subfigure}
  \begin{subfigure}{0.136\linewidth}
  	\includegraphics[width=\textwidth]{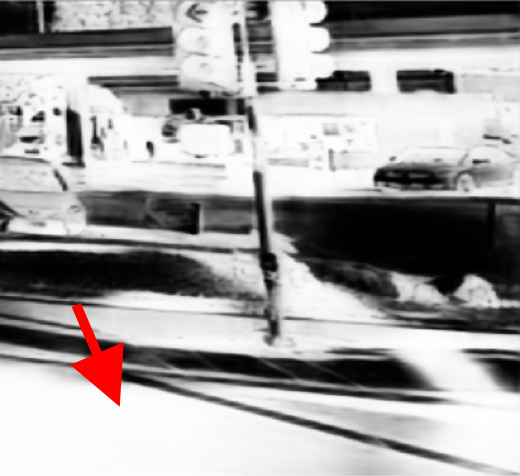}
  \end{subfigure}

  \ \\
  
  \vspace*{0.3mm}
  \begin{subfigure}{0.136\linewidth}
  	\includegraphics[width=\textwidth]{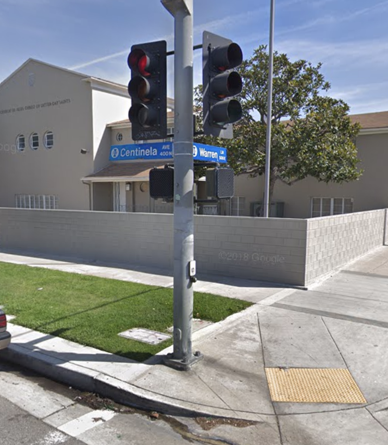}
  	\vspace{-5.5mm} \caption{Input image}
  \end{subfigure}
  \begin{subfigure}{0.136\linewidth}
  	\includegraphics[width=\textwidth]{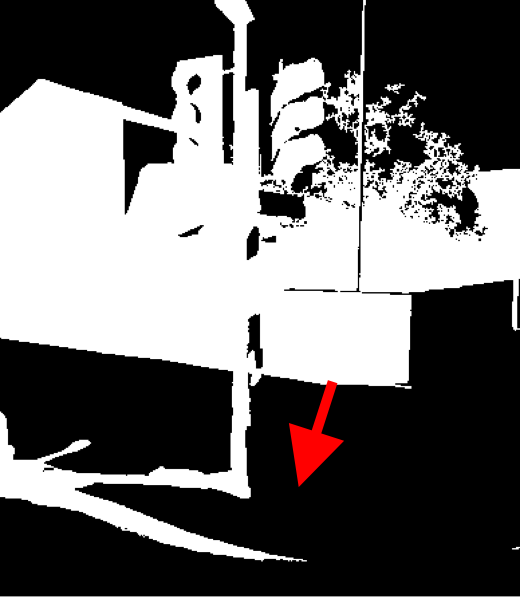}
  	\vspace{-5.5mm} \caption{Ground truth}
  \end{subfigure}
  \begin{subfigure}{0.136\linewidth}
  	\includegraphics[width=\textwidth]{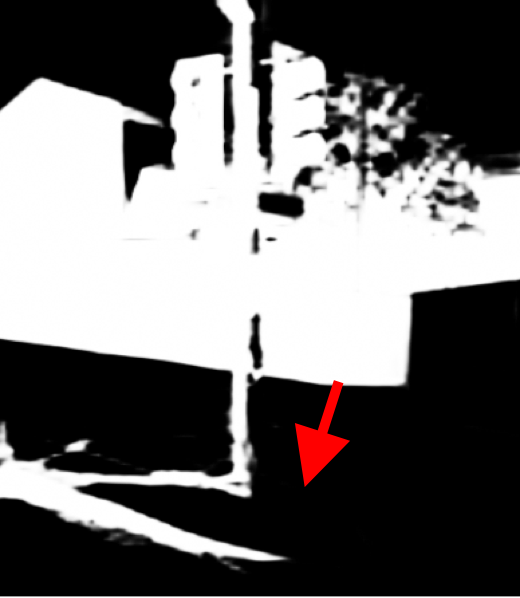}
  	\vspace{-5.5mm} \caption{FSDNet (ours)}
  \end{subfigure}
  \begin{subfigure}{0.136\linewidth}
  	\includegraphics[width=\textwidth]{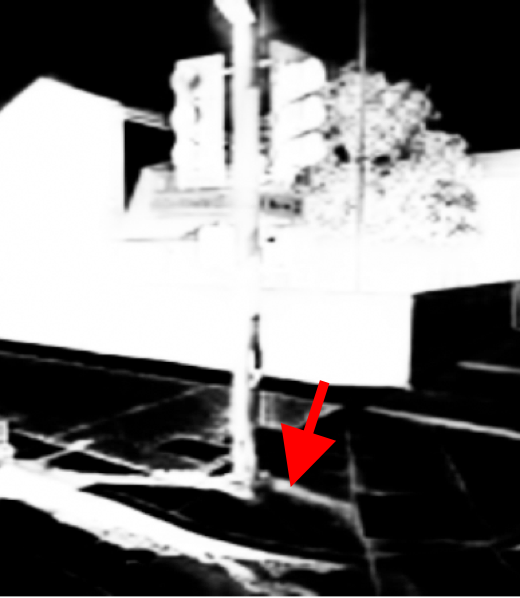}
  	\vspace{-5.5mm} \caption{DSDNet~\cite{zheng2019distraction}}
  \end{subfigure}
  \begin{subfigure}{0.136\linewidth}
  	\includegraphics[width=\textwidth]{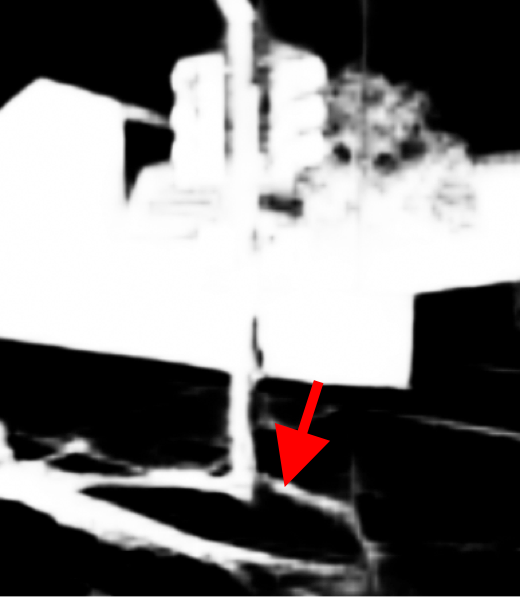}
    \vspace{-5.5mm} \caption{BDRAR~\cite{zhu2018bidirectional}}
  \end{subfigure}
  \begin{subfigure}{0.136\linewidth}
  	\includegraphics[width=\textwidth]{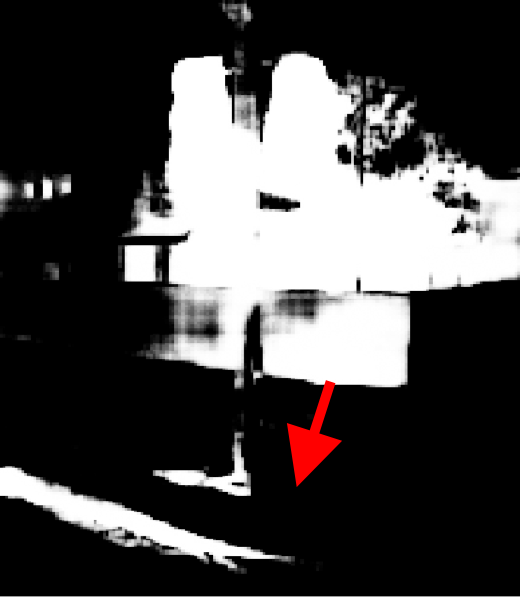}
  	\vspace{-5.5mm} \caption{A+D Net~\cite{le2018a+d}}
  \end{subfigure}
  \begin{subfigure}{0.136\linewidth}
  	\includegraphics[width=\textwidth]{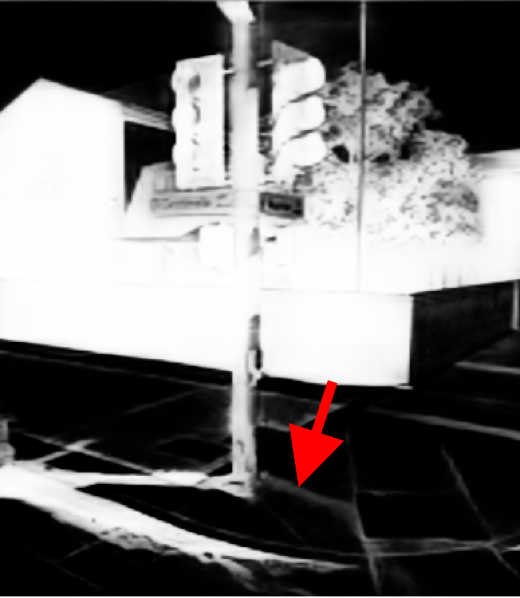}
  	\vspace{-5.5mm} \caption{DSC~\cite{hu2019direction,Hu_2018_CVPR}}
  \end{subfigure}
  
	\caption{More visual comparison results (continue from Figure~\ref{fig:comparisons_visual}).}
	\label{fig:comparisons_visual2}
\end{figure*}


\subsection{Evaluation on the Network Design}

We evaluate the effectiveness of the major components in FSDNet using the validation set of our data.
First, we build a ``basic'' model using the last layer of the backbone network
to directly predict shadows.
This model is built by removing the DSC module, DEM, and skip connections of the low- and middle-level features in the architecture shown in Figure~\ref{fig:architecture}.
Then, we add back the DSC module to aggregate global features; the model is ``basic+DSC'' in Table~\ref{table:ablation}.
Further, we consider the low-level features refined by the DEM and set up another network model, namely ``FSDNet w/o DEM,'' by removing the DEM from the whole architecture and directly concatenating the low-, middle-, and high-level features.
\revised{To evaluate the effectiveness of the multi-scale features, we further build ``FSDNet-high-only'' by removing the multi-scale features and adopting the high-level feature map to directly predict the shadow masks.}
From the quantitative evaluation results shown in Table~\ref{table:ablation}, we can see that the major components help improve the results and contribute to the full pipeline.
%

%

\subsection{Comparing with the State-of-the-art}
\label{sec:experiments_SOTA}

\noindent
\textbf{Comparison with recent shadow detection methods.} \ 
We consider four recent shadow detection methods: DSDNet~\cite{zheng2019distraction}, BDRAR~\cite{zhu2018bidirectional}, A+D Net~\cite{le2018a+d}, and DSC~\cite{hu2019direction,Hu_2018_CVPR}.
We re-train each of the models on the training set of all categories in our dataset and evaluate them on our testing set.
For a fair comparison, we resize all input images to be 512$\times$512.

Table~\ref{table:compare_with_SOTA} reports the overall quantitative comparison results, from which we observe:
(i) Our network uses only 4M parameters and is able to process 77.52 frames per second (see the second column on FPS in Table~\ref{table:compare_with_SOTA}) on a single GeForce GTX 1080 Ti GPU.
Particularly, it performs faster and achieves more accurate results than the recent real-time shadow detector A+D Net.
(ii) Our method performs favorably against all the other methods on the five categories of scenes in our dataset in terms of $F_\beta^\omega$ and achieves comparable performance with DSDNet~\cite{zheng2019distraction} on BER.  
(iii) DSDNet requires the results of BDRAR~\cite{zhu2018bidirectional}, A+D Net~\cite{le2018a+d}, and DSC~\cite{hu2019direction,Hu_2018_CVPR} to discover false positives and false negatives and to train its model during the training.
Hence, the entire training time is a summation of the time of the four methods.
In contrast, we train our FSDNet using only the ground truths and the training time is only $3.75$ hours on a single GeForce GTX 1080 Ti GPU.
(iv) Our method performs better on $F_\beta^\omega$ than BER.
This is because BER is designed for evaluating binary predictions, so predicted values that are smaller than $0.5$ are all set as zeros, and ones, otherwise.
In contrast, $F_\beta^\omega$ is designed for evaluating continuous predictions. 
Values predicted by FSDNet tend to near zero or one, while DSDNet and BDRAR tend to produce many gray values; see the false positives in the last two rows of Figure~\ref{fig:comparisons_visual} and the false negatives in the first row of Figure~\ref{fig:comparisons_visual2}. 
Unlike $F_\beta^\omega$, BER cannot account for these false predictions.



\begin{table*}  [tp]
	\begin{center}
		\caption{\revised{Category analysis on our dataset. The results are evaluated on the validation set.}}
		\label{table:category}
  	\resizebox{0.98\linewidth}{!}{%
			\begin{tabular}{c|c|c|c|c|c|c|c|c|c|c|c|c}
				\hline
			    \multirow{2}{*}{Training Set} & \multicolumn{2}{c|}{Overall} & \multicolumn{2}{c|}{shadow-ADE}& \multicolumn{2}{c|}{shadow-KITTI} & \multicolumn{2}{c|}{shadow-MAP} & \multicolumn{2}{c|}{shadow-USR} & \multicolumn{2}{c}{shadow-WEB}  \\
			    
			    \cline{2-13}
			    & $F_\beta^\omega$ & BER & $F_\beta^\omega$ & BER & $F_\beta^\omega$ & BER & $F_\beta^\omega$ & BER & $F_\beta^\omega$ & BER & $F_\beta^\omega$ & BER \\
	
				\hline
				
				Shadow-ADE (793 images) & 74.73 & \textbf{12.98} & \textbf{75.24} & \textbf{13.11} & 80.43 & 13.35 & 76.13 & 14.67 & 66.39 & 12.64 & 75.45 & 13.23 \\
				Shadow-KITTI (1941 images) & 67.95 & 19.91 & 58.37 & 26.51 & \textbf{88.28} & \textbf{8.31} & 64.12 & 23.91 & 52.76 & 26.97 & 67.12 & 19.88  \\
				Shadow-MAP (1116 images) & 77.24 & 13.19 &  72.90 & 15.23 & 82.71 & 12.06 & \textbf{84.38} & \textbf{9.70} & 69.16 & 15.74 & 76.53 & 13.77 \\
				Shadow-USR (1711 images) & 75.50 & 21.33 & 59.67 & 27.45 & 77.54 & 17.28 & 62.26 & 25.22 & \textbf{89.01} & \textbf{5.21} & 75.60 & 16.03 \\
				Shadow-WEB (1789 images) & \textbf{79.91} & 13.37 & 73.48 & 15.67 & 80.41 & 13.79 & 78.68 & 13.47 & 79.19 & 9.19 & \textbf{83.67} & \textbf{9.74} \\
			
				\hline
				
			\end{tabular} }
	\end{center}
	\vspace{-2mm}
\end{table*}

%
%

%

Further, Figures~\ref{fig:comparisons_visual} \&~\ref{fig:comparisons_visual2} show visual comparison results.
Our results are more consistent with the ground truths, while other methods may mis-recognize black regions as shadows,~\eg, the green door in the first row and the trees in the last three rows of Figure~\ref{fig:comparisons_visual} and the last two rows of Figure~\ref{fig:comparisons_visual2}, or fail to find unobvious shadows,~\eg, the shadow across backgrounds of different colors in the first row of Figure~\ref{fig:comparisons_visual2}. 
%
However, our method may also fail to detect some extremely tiny shadows of the trees and buildings; see Figure~\ref{fig:comparisons_visual} and the last two rows of Figure~\ref{fig:comparisons_visual2}.
In the future, we plan to further enhance the DEM by considering patch-based methods to process the image regions with detailed structures in high resolutions.

\vspace{1.5mm}
\noindent
\textbf{Comparison with other networks.} \ 
%
Deep network architectures for saliency detection, mirror detection, and semantic segmentation may also be used for shadow detection, if we re-train their models on shadow detection datasets.
We took three recent works on saliency detection (\ie, $R^{3}$Net~\cite{deng18r}), mirror detection (\ie, MirrorNet~\cite{Yang_2019_ICCV}), and semantic segmentation (\ie, PSPNet~\cite{zhao2017pyramid}), re-trained their models on our training set, then evaluated them on our testing set.
The last three rows in Table~\ref{table:compare_with_SOTA} show their quantitative results.
Comparing our results with theirs, our method still performs favorably against these deep models for both accuracy and speed.

\subsection{\revised{Category Analysis on Our Dataset}}

\revised{To explore the domain differences between various categories of shadow images in our CUHK-Shadow dataset, we performed another experiment by training a deep model on training images from each data category and then evaluating the trained models on images of different categories in the validation set. From the results reported in Table~\ref{table:category}, we can see that the trained models achieve the best performance on their corresponding data categories, and the performance may degrade largely on other categories. These results show that there exist large domain differences between shadow images collected from different scenarios.}


\subsection{Application} 
\begin{figure}[t]
  	\centering
\begin{subfigure}{0.47\linewidth}
	\includegraphics[width=\textwidth]{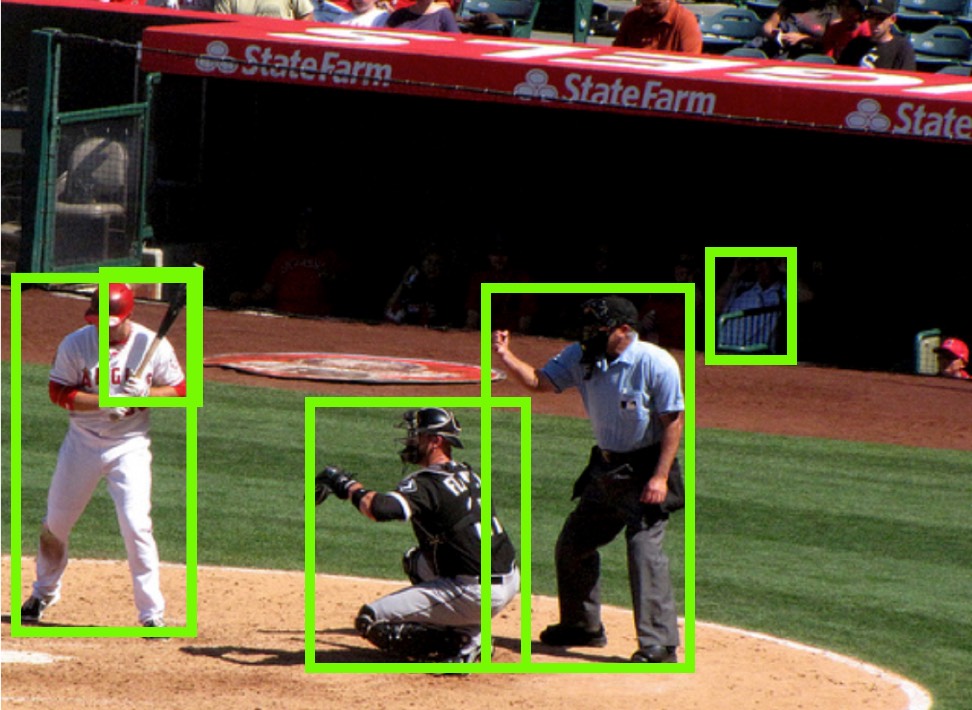}
	\vspace{-5.5mm}
	\caption{\footnotesize{Original image}}
	
\end{subfigure}
\begin{subfigure}{0.47\linewidth}
	\includegraphics[width=\textwidth]{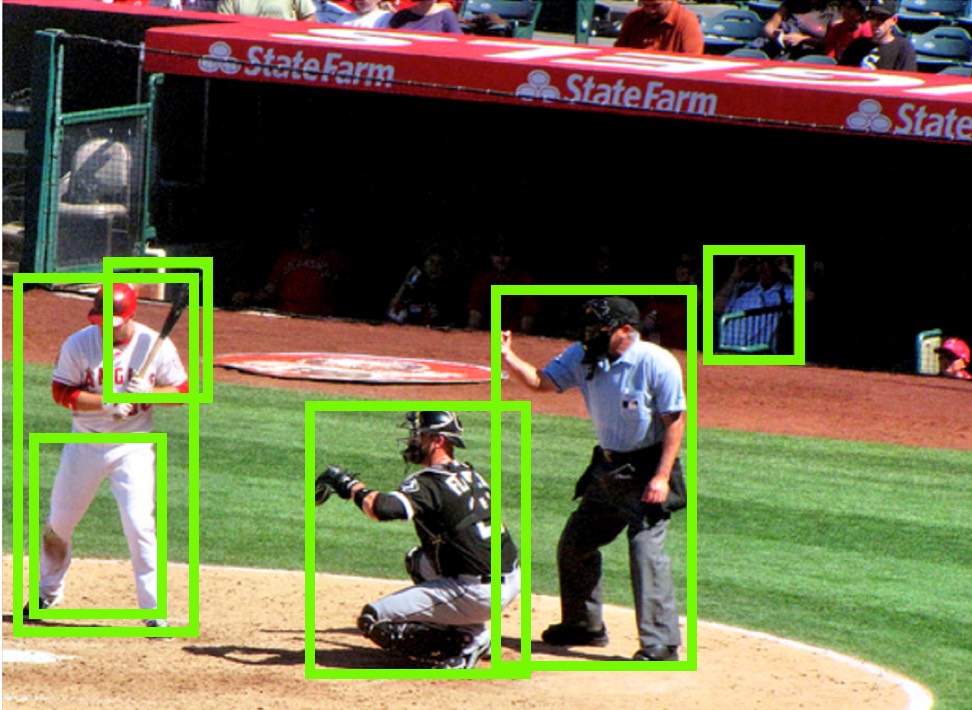}
	\vspace{-5.5mm}
	\caption{\footnotesize{Result of DeepUPE}~\cite{wang2019underexposed}}
	
\end{subfigure}
\ \\

\vspace*{1mm}
\begin{subfigure}{0.47\linewidth}
	\includegraphics[width=\textwidth]{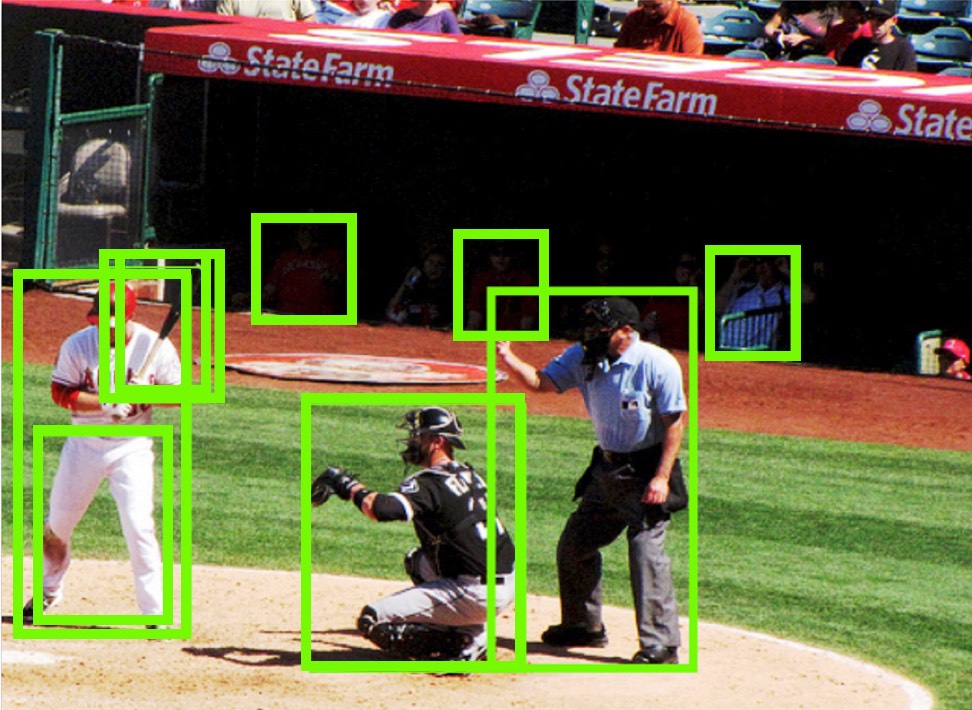}
	\vspace{-5.5mm}
	\caption{\centering \footnotesize{Histogram equalization	over \newline the whole image}}
	
\end{subfigure}
\begin{subfigure}{0.47\linewidth}
	\includegraphics[width=\textwidth]{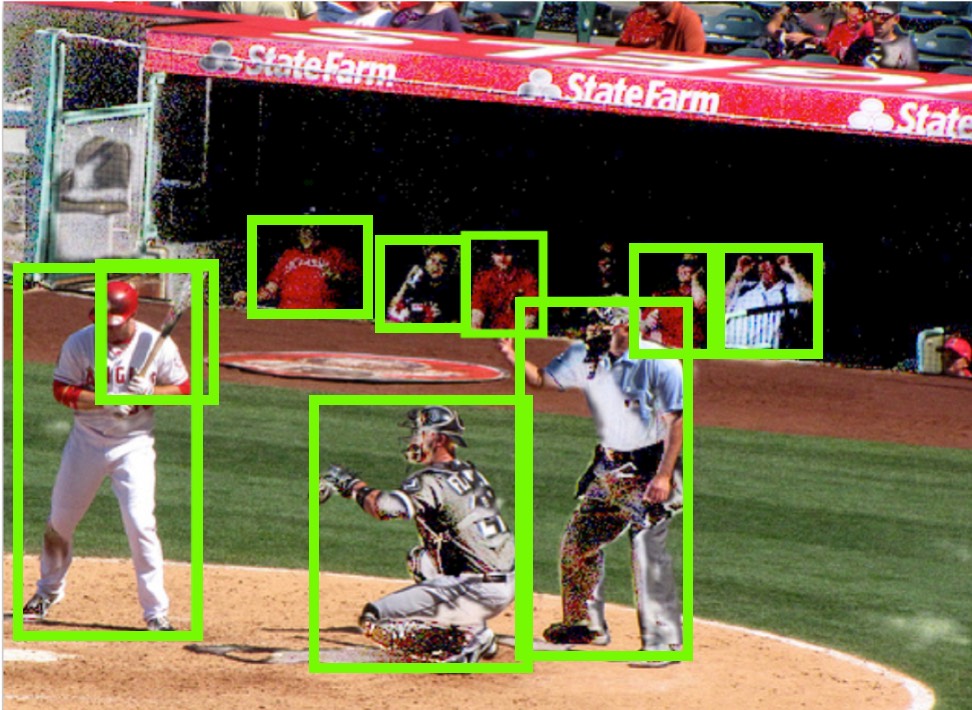}
	\vspace{-5.5mm}
	\caption{\centering \footnotesize{Histogram equalization over \newline the shadow regions}}
	
\end{subfigure}

	\caption{Object detection results on different inputs.}
	\label{fig:application}
    \vspace{-2mm} 
\end{figure}

%
%
Figure~\ref{fig:application} shows an example of how shadow detection helps object detection in certain situations.
The objects are detected by Google Cloud Vision\footnote{https://cloud.google.com/vision/}, but the people under the grandstand were missed out, due to the presence of the self shadows; see Figure~\ref{fig:application} (a).
%
After adopting a recent underexposed photo enhancement method,~\ie, DeepUPE~\cite{wang2019underexposed} or a simple histogram equalization operation, to adjust the contrast over the whole image, we can enhance the input image.
However, the improvement on the object detection is still limited; see Figures~\ref{fig:application}~(b) \& (c).
If we apply the histogram equalization operation only on the shadows with the help of our detected shadow mask, we can largely improve the visibility of the people in the shadow regions and also improve the object detection performance, as demonstrated in Figure~\ref{fig:application}~(d).
\revised{However, the changed appearance caused by histogram equalization may hurt the performance of other cases.
In the future, we will explore a shadow-mask-guided method for photo enhancement to improve both the visual quality and performance of high-level computer vision tasks.}



%




\section{Conclusion}
\label{sec:conclusion}

This paper revisits the problem of shadow detection, with a specific aim to handle general real-world situations.
We collected and prepared a new benchmark dataset of 10,500 shadow images, each with a labeled shadow mask, from five different categories of sources.
Comparing with the existing shadow detection datasets, our CUHK-Shadow dataset comprises images for diverse scene categories, features both cast and self shadows, and also introduces a validation set to reduce the risk of overfitting.
We show the complexity of CUHK-Shadow by analyzing the shadow area proportion, number of shadows per image, shadow location distribution, and color contrast between shadow and non-shadow regions.
Moreover, we design FSDNet, a new and fast deep neural network architecture, and formulate the detail enhancement module to bring in more shadow details from the low-level features.
Comparing with the state-of-the-arts, our network performs favorably for both accuracy and speed.
From the results, we can see that all methods (including ours) cannot achieve high performance on our new dataset, different from what have been achieved in the existing datasets.
In the future, we plan to explore the performance of shadow detection in different image categories, to prepare shadow data for indoor scenes, and to provide non-binary masks to indicate soft shadows.


\section*{Acknowledgment}
This work was supported by a grant from the Key-Area Research and Development Program of Guangdong Province, China (2020B010165004), a grant from the Research Grants Council of the Hong Kong Special Administrative Region, China (Grant No. CUHK 14201620), and a grant from the CUHK Direct Grant for Research.


%

{\small
	\bibliographystyle{IEEEtran}
	\bibliography{reference}
}


\ifCLASSOPTIONcaptionsoff
  \newpage
\fi

\vspace*{-7mm}

\begin{IEEEbiography}[{\includegraphics[width=1in,height=1.25in,clip,keepaspectratio]{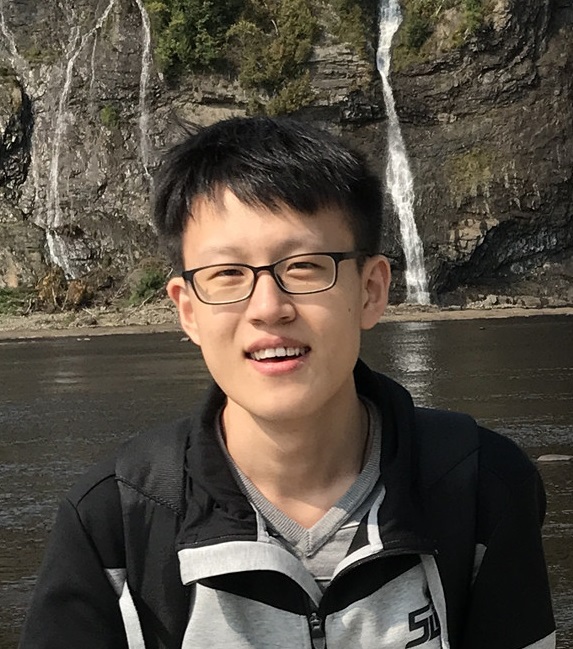}}]{Xiaowei Hu}
received his Ph.D. degree in Computer Science and Engineering from the Chinese University of Hong Kong in 2020, and his B.Eng. degree in Computer Science and Technology from South China University of Technology in 2016. He is currently working as a postdoctoral fellow in the Chinese University of Hong Kong. His research interests include computer vision, deep learning, and low-level vision.
\end{IEEEbiography}

\vspace*{-7mm}
\begin{IEEEbiography}[{\includegraphics[width=1in,height=1.25in,clip,keepaspectratio]{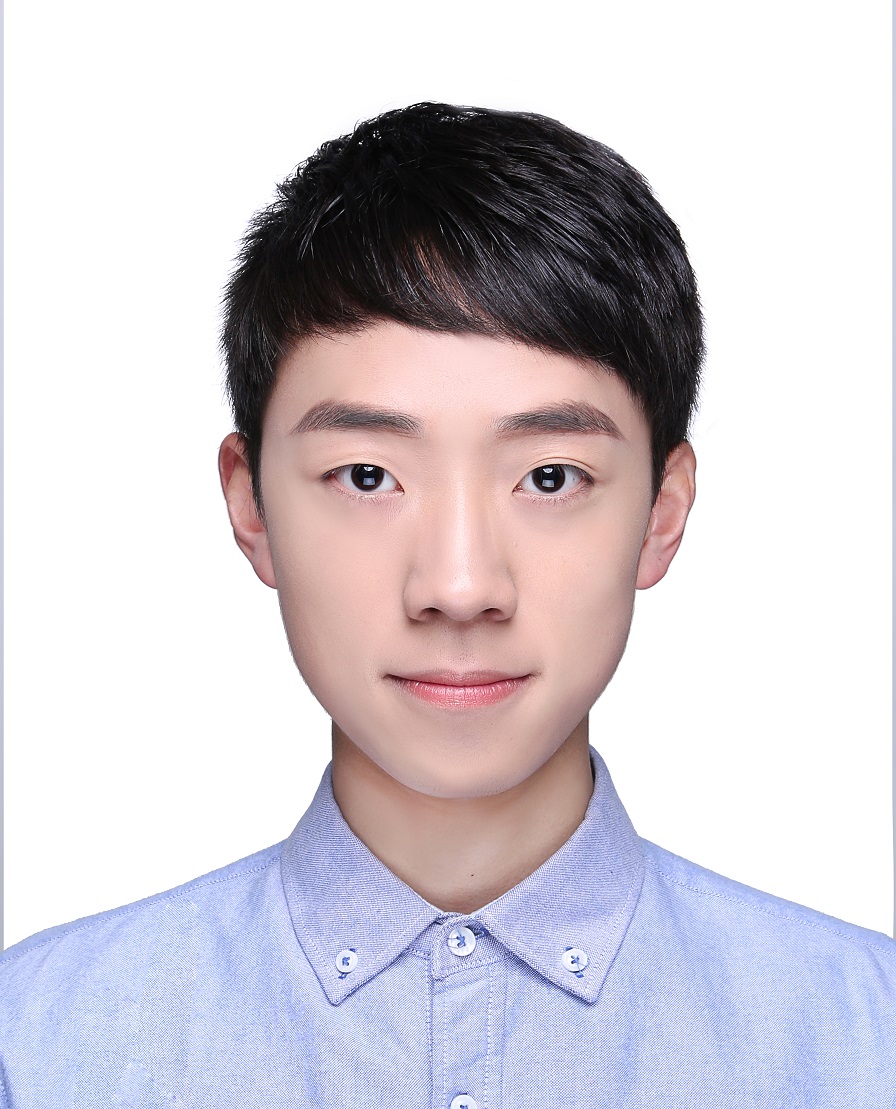}}]{Tianyu Wang} received his B.Eng. degree in Computer Science and Technology from Dalian University of Technology, China, in 2018.  He is currently working toward his Ph.D. degree in the Department of Computer Science and Engineering, The Chinese University of Hong Kong. His research interests include computer vision, image processing, computational photography, low-level vision, and deep learning.
\end{IEEEbiography}

\vspace*{-7mm}
\begin{IEEEbiography}[{\includegraphics[width=1.0in,height=1.25in,clip,keepaspectratio]{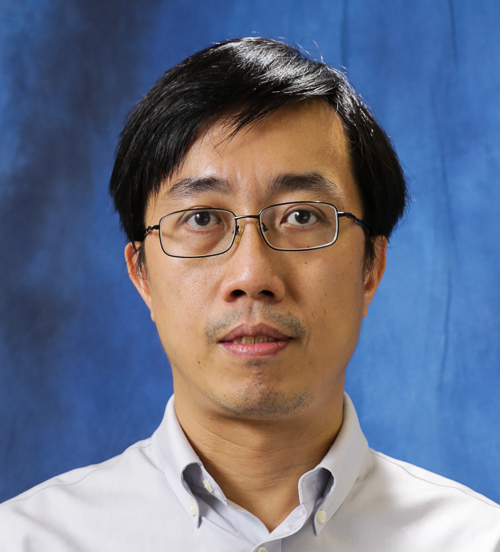}}]{Chi-Wing Fu} is currently an associate professor in the Chinese University of Hong Kong.  He served as the co-chair of SIGGRAPH ASIA 2016's Technical Brief and Poster program, associate editor of IEEE Computer Graphics \& Applications and Computer Graphics Forum, panel member in SIGGRAPH 2019 Doctoral Consortium, and program committee members in various research conferences, including SIGGRAPH Asia Technical Brief, SIGGRAPH Asia Emerging tech., IEEE visualization, CVPR, IEEE VR, VRST, Pacific Graphics, GMP, etc.  His recent research interests include computation fabrication, point cloud processing, 3D computer vision, user interaction, and data visualization.
\end{IEEEbiography}


\vspace*{-7mm}
\begin{IEEEbiography}[{\includegraphics[width=1in,height=1.25in,clip,keepaspectratio]{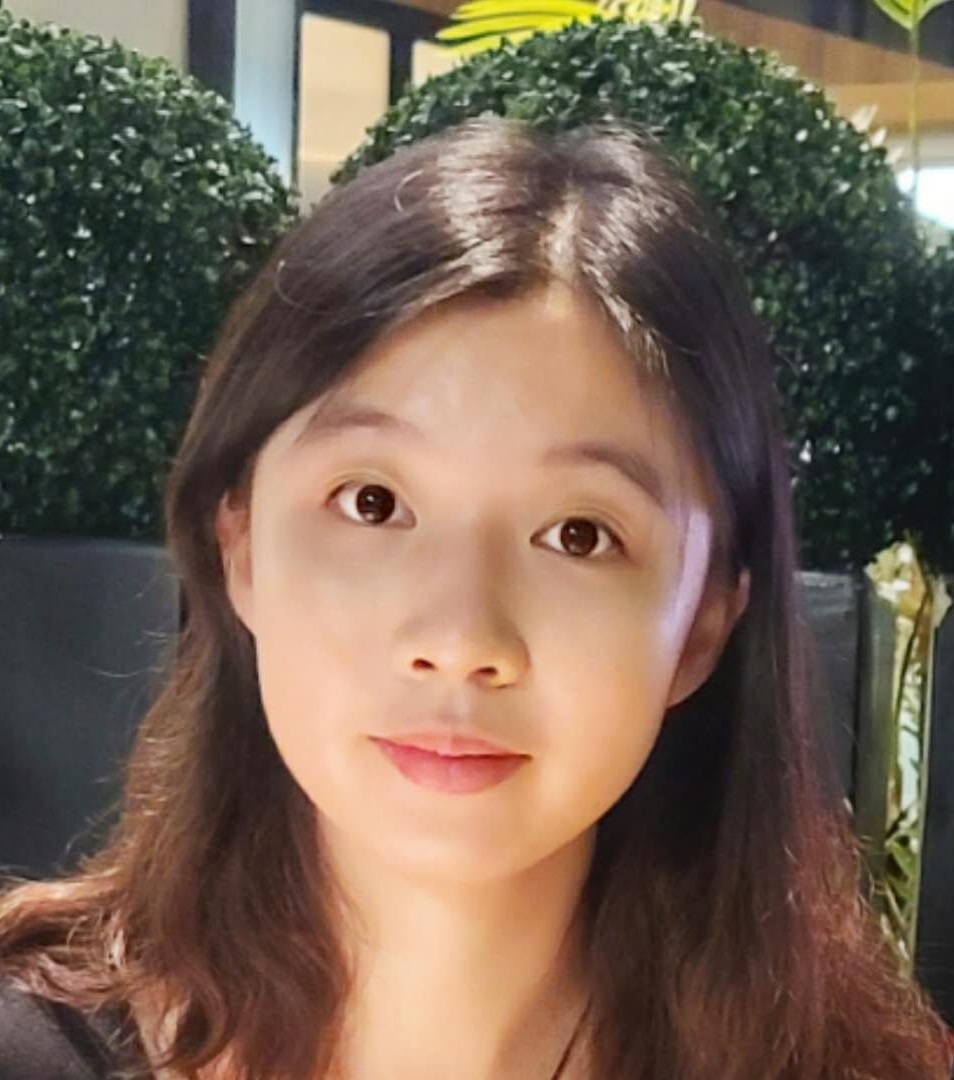}}]{Yitong Jiang} is pursuing her B.Sc. degree (Computer Science) in the Chinese University of Hong Kong since 2016. Her research interests include computer vision, deep learning, image processing, and computational photograph.
\end{IEEEbiography}

\vspace*{-7mm}
\begin{IEEEbiography}[{\includegraphics[width=1in,height=1.25in,clip,keepaspectratio]{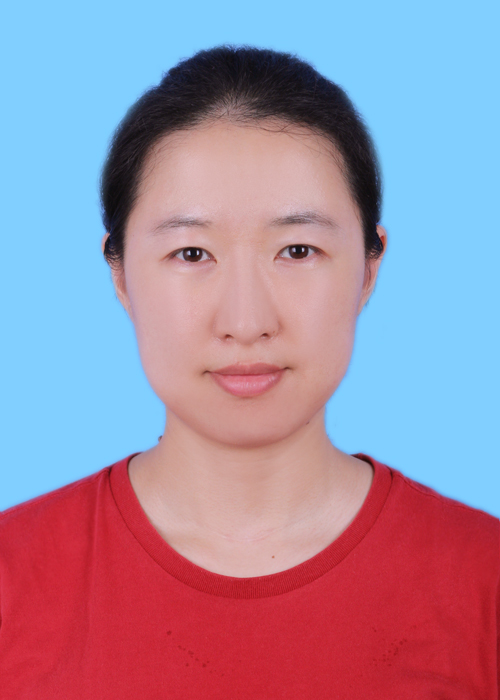}}]{Qiong Wang} is an Associate Researcher with the Shenzhen Institute of Advanced Technology, Chinese Academy of Sciences, Shenzhen, China. Her research interests include VR applications in medicine, visualization, medical imaging, human–computer interaction, and computer graphics.
\end{IEEEbiography}

\vspace*{-7mm}
\begin{IEEEbiography}[{\includegraphics[width=1in,height=1.25in,clip,keepaspectratio]{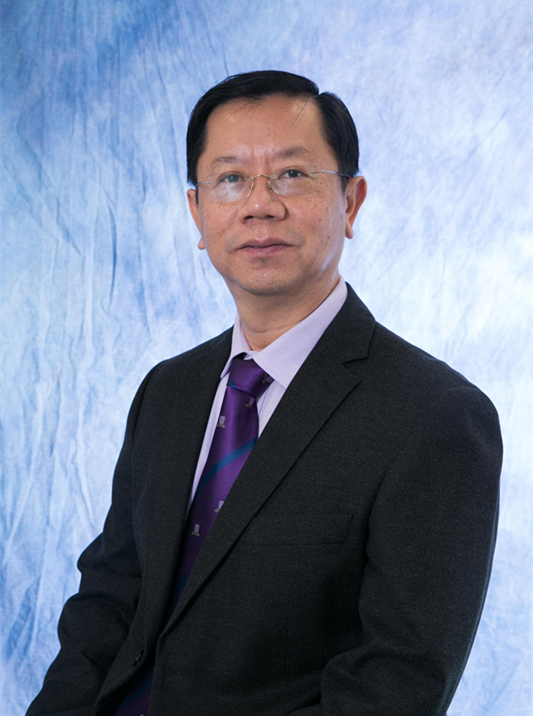}}]{Pheng-Ann Heng} received his B.Sc. (Computer Science) from the National University of Singapore in 1985. 
	He received his M.Sc. (Computer Science), M. Art (Applied Math) and Ph.D. (Computer Science) all from the Indiana University in 1987, 1988, 1992 respectively.
	He is a professor at the Department of Computer Science and Engineering at The Chinese University of Hong Kong. He has served as the Department Chairman from 2014 to 2017 and as the Head of Graduate Division from 2005 to 2008 and then again from 2011 to 2016.
	He has served as the Director of Virtual Reality, Visualization and Imaging Research Center at CUHK since 1999. He has served as the Director of Center for Human-Computer Interaction at Shenzhen Institutes of Advanced Technology, Chinese Academy of Sciences since 2006. He has been appointed by China Ministry of Education as a Cheung Kong Scholar Chair Professor in 2007. 
	His research interests include AI and VR for medical applications, surgical simulation, visualization, graphics, and human-computer interaction.
\end{IEEEbiography}

\end{document}